%% file: main_arXiv.tex
\documentclass[11pt, lefttitle]{behroozarxiv}

\input{preamble.tex}

\title{Human--AI Co-Interpretation for Responsible AI:\\
A Hermeneutic Perspective%
\thanks{The author's primary training is in computer science. This work arises from an interdisciplinary interest in hermeneutic philosophy rather than formal philosophical schooling. Consequently, the hermeneutic concepts are applied from the perspective of a computer science researcher exploring philosophy.}%
}

\DeclareRobustCommand{\orcidicon}{%
  \href{https://orcid.org/0000-0001-9568-4166}{%
    \raisebox{-0.2ex}{\includegraphics[height=1.6ex]{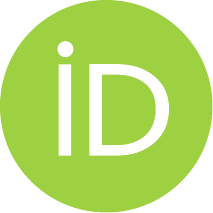}}%
  }%
}

\author{Behrooz Razeghi~\orcidicon}
\affiliation{School of Engineering and Applied Sciences, Harvard University}

\abstract{
Across law, education, policy analysis, and public moral argumentation, LLM outputs are being used often for work that requires interpretations to be justified with textual evidence and explicit normative standards. Yet a recurrent failure mode -- what I call \textit{interpretive misplacement} -- is that model-generated readings get treated as settled meanings without an explicit interpretive frame (sources, scope constraints, normative commitments), without preserving defensible alternatives, and without provenance that lets readers find the supporting passages. In such settings, the risk is not only factual error but lost accountability: readers and institutions cannot reliably assess what an output commits them to, or on what basis. Drawing on philosophical hermeneutics, this paper discusses this risk and derives design principles for structuring human-AI co-interpretation. The paper also provides a structured synthesis of recent scholarship on hermeneutics and AI, organizing this emerging literature into a set of recurrent lines of argument and design-relevant gaps. LLM outputs are treated as candidate readings, whereas hermeneutic understanding is reserved for accountable human interpreters situated in disciplinary historical-linguistic traditions. Human-AI interaction is characterized as an AI-mediated interpretive loop. Hermeneutic understanding is distinguished from token-prediction--based text generation. On this basis, existing LLM techniques are reorganized into design patterns for hermeneutically responsible use in interpretive settings. Finally, the discussion turns to implications for legal practice, educational assessment and feedback, scholarly knowledge production, and public moral argumentation. It also treats digital hermeneutics as a literacy: the capacity to read AI-mediated texts by examining frames, provenance, and readings, and by contesting outputs.
}

\preprint{arXiv preprint}
\date{\today}
\begin{document}

\maketitle

\clearpage

{
% \small
  \hypersetup{linkcolor=BehroozNavy}
  \tableofcontents
}

\clearpage

\section{Introduction}

\vspace{-4pt}

\subsection{Motivation}

\vspace{-2pt}

In courts, classrooms, policy work, and public moral argumentation, LLM outputs are increasingly treated as reasons---as readings that must be normatively or contextually justified to relevant audiences---not simply as summaries of retrieved facts. In these settings, interpretive safety failures can arise even when an output is fluent and locally plausible: coherent at the sentence level yet misframed at the level of reasons, warrants, and justification. The risk is therefore not only factual error but \textit{interpretive misplacement}: treating fluent model output as determinate meaning outside an explicitly declared interpretive frame (sources, scope constraints, normative commitments), without presenting competing defensible readings, and without evidential support and traceability. This dynamic can produce an \textit{illusion of understanding}, in which users over-accept fluent outputs and under-exercise interpretive responsibility---for example, by not checking supporting passages, not soliciting alternative readings, and not contesting scope or value assumptions. In research on human--automation interaction, this failure mode corresponds to misuse and overreliance on automation \citep{parasuraman1997humans} and to automation bias, including commission errors (accepting incorrect automated advice) and omission errors (failing to act or to check when automation does not cue appropriately) \citep{skitka1999does}. Recent work argues that LLMs can produce illusions of understanding in knowledge practices, and that these illusions may encourage over-acceptance while discouraging verification \citep{messeri2024artificial, marchetti2025artificial}. Yet comparatively little work treats the interpretive interaction itself \citep{gonzalez2024beyond, delacroix2025designing, delacroix2025moral, stanczak2025societal}---how users adopt, contest, and authorize outputs---as a central target for design and evaluation.

\begin{figure}[t!]
\centering
\includegraphics[width=\linewidth]{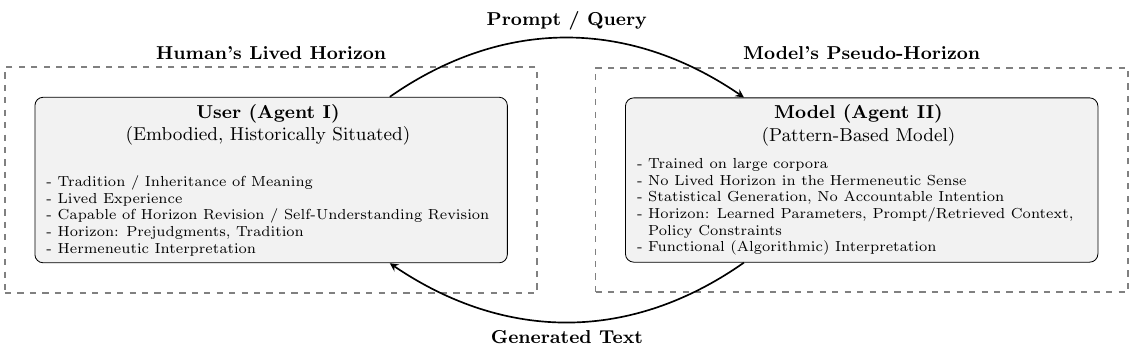}
\caption{Conceptual schematic of the human--AI interpretive loop (AI-mediated interpretive process). 
Agent~I (a historically situated human) formulates a prompt from within a lived horizon shaped by pre-understanding. Agent~II (a pattern-based LLM system) generates a continuation through algorithmic transformation of the prompt (and any retrieved context), conditioned by learned parameters and system constraints rather than a lived horizon. The human then appropriates and recontextualizes the output within that horizon, potentially revising understanding and reframing subsequent prompts.}
\label{fig:hermeneutic-llm-relationship}
\vspace{5pt}
\end{figure}

This paper discusses an interaction-centered hermeneutic perspective on human--AI co-interpretation. It treats LLM outputs as \textit{interpretive proposals}, i.e., statistical continuations produced under training- and prompt-level constraints that become action-guiding only through human appropriation (interpretation and use). To structure the discussion, I use the notion of an \textit{AI-mediated interpretive loop}: the human formulates a question from within a situated horizon; the model responds under constraints set by training data, retrieval, and prompts; and the human evaluates, revises, and recontextualizes that response.
The high-level structure of this AI-mediated interpretive loop is shown in Figure~\ref{fig:hermeneutic-llm-relationship}; a more detailed operational view is provided in Figure~\ref{fig:two-agent-interpretive-loop}.
This framing shifts attention from whether the model has understanding to which interactional conditions support accountable interpretation---that is, when and why users treat outputs as authoritative readings, and how interpretive error is produced in practice. The contribution is conceptual and design-oriented: it clarifies the structure of interpretive trust and proposes implementable ways to make justification, scope, and provenance inspectable.

The conceptual contribution is a disciplined distinction between \textit{hermeneutic understanding} and \textit{artificial (algorithmic) interpretation}, understood here as token-prediction--based text generation that yields candidate readings without historicity or accountability. Building on modern hermeneutics, I reserve understanding for historically situated interpreters and describe model behavior as a transformation of prompts and corpora into textual candidates. The practical challenge is to clarify what safe use requires in interpretive domains without implying model agency. I therefore discuss requirements for system and interface design that can be implemented with current prompting, retrieval, and workflow constraints. In the LLM-mediated interpretive settings considered here, this safety-motivated stance entails three recurring commitments: (i) retain human responsibility for interpretive judgment where justification is required; (ii) treat collaboration as a target capability rather than an incidental interface feature; and (iii) use users' ability to direct, constrain, and contest outputs as a safeguard when meaning and judgment are at stake.
These commitments motivate the system-level and pedagogical proposals that follow.

Guided by this premise, I (i) articulate the AI-mediated interpretive loop as a vocabulary for describing LLM use in interpretive work; (ii) reorganize existing capabilities into a design language for hermeneutically responsible AI-mediated interpretation, including artificial horizons (context headers, ontologies, jurisdictional scopes), system-level normative constraints (institutional or cultural policies), generated meta-commentary and counter-questions, and contrastive outputs; (iii) propose \textit{hermeneutic quality} as an evaluation agenda that complements task accuracy with part--whole integrity, the identification and preservation of ambiguity, plurality (the presentation of multiple defensible readings), evidential support and traceability, limitation-signaling (scope, uncertainty, alternatives), and user appropriation (uptake in the user's own deliberation and action); and (iv) articulate \textit{digital hermeneutics} as a human-facing literacy for interpreting and contesting AI-mediated texts in professional and educational settings.

For a related argument in the more specific domain of AI alignment, see \cite{razeghi2026principles}, which argues that principle-specified alignment includes an interpretive component because general principles do not determine their own application in concrete cases.

\subsection{Position and Contributions}

I take a practice-centered view of LLMs and hermeneutics. The central question here is not whether models \textit{understand}, but how to organize human--AI interaction when LLMs function as instruments within interpretive workflows---so that AI-mediated interpretation remains plural, accountable, and traceable.
I reserve \textit{hermeneutic understanding} for historically situated human interpreters, and I describe model behavior as \textit{artificial (algorithmic) interpretation}. This shifts the emphasis to co-interpretation itself: how it should be designed, evaluated, and taught so that human horizons stay explicit, ambiguity and disagreement are not smoothed over, and interpretive claims remain traceable to sources and to the situated contexts in which they are made.
This paper advances a conceptual argument with practice-oriented implications, rather than introducing new empirical benchmarks. Its contributions fall along three axes:

\newpage

\begin{itemize}[leftmargin=*]
    \item 
    \textbf{With respect to existing scholarship:}
    \begin{itemize}[leftmargin=*]
        \item[(A)]
        \textbf{An overview of recent work on hermeneutics and AI.}
        I provide a structured review of recent scholarship that explicitly brings philosophical hermeneutics into conversation with LLMs. From this material, I build a map of the different claims being made about understanding,'' meaning,'' and ``context'' in LLM discourse, tracking genuine points of convergence while also flagging systematic cross-talk, where authors rely on incompatible senses of understanding. That synthesis makes the remaining conceptual gaps and evaluation desiderata hard to miss, and it motivates the need for a co-interpretation framework.
        \item[(B)] 
        \textbf{Interaction-centered reframing of hermeneutics--AI debates.}
        I reconstruct key strands of modern hermeneutics with an eye to how they characterize understanding as historical, dialogical, and structurally circular. I then use this theoretical lens to organize and assess emerging work on hermeneutics and AI, including recent engagements with LLMs. I argue that these contributions largely remain \textit{model-centered} (do LLMs understand?) or \textit{text-centered} (how should we read AI texts?). 
        I shift the focus to an \textit{interaction-centered} question: what counts as more or less responsible human–AI co-interpretation in situated interpretive practices? I clarify how my proposal differs by schematizing human--model interaction as an AI-mediated interpretive loop and by introducing \textit{hermeneutic quality} as a way to talk about better or worse uses of LLMs in interpretive domains without attributing understanding to the models.
        \item[(C)]
        \textbf{Conceptual clarification: hermeneutic understanding vs artificial interpretation.}
        Building on Schleiermacher, Dilthey, Heidegger, Gadamer, Ricœur and Derrida, I propose a terminological distinction between \textit{hermeneutic understanding} and \textit{artificial (algorithmic) interpretation}. I reserve the former for historically situated human interpreters and use the latter for pattern-based model behavior. 
        On this basis, I argue that LLMs are better described as \textit{proposal-generating instruments} within a two-stage interpretive loop (model generation followed by human appropriation), and that normative assessment should target the quality of this co-interpretive loop rather than attributing quasi personhood to models.
    \end{itemize}
    \item 
    \textbf{With respect to existing practice:}
    \begin{itemize}[leftmargin=*]
        \item[(D)] 
        \textbf{A conceptual design language for hermeneutic human--AI co‑interpretation.}
        Current LLMs already support a set of capabilities that can be repurposed for hermeneutic purposes: perspectival prompts (for example, answering as a legal historian), multiple interpretations on request, multi-turn dialogue, retrieval-augmented answers, and constrained generation. 
        I treat these capabilities as a conceptual design language for what I call \textit{hermeneutic AI}. It includes \textit{artificial horizons} (context headers, domain ontologies, jurisdictional and doctrinal scopes, institutional value statements) that act as pseudo-traditions. It also includes \textit{cultural and ethical overlays} that make normative starting points explicit, as well as auto-commentary and self-questioning patterns that expose uncertainty and alternative readings. Finally, \textit{contrastive outputs} (parallel, labeled interpretations) make disagreement and alternatives explicit.
        All of these patterns are expressible with today's prompt-engineering, retrieval, and workflow and interface constraints.
        \item[(E)] 
        \textbf{Hermeneutic quality as an evaluation agenda.}
        I introduce \textit{hermeneutic quality} as an evaluative agenda for AI-mediated interpretation, and I organize existing techniques into six criteria: part--whole integrity, ambiguity sensitivity, plurality, evidence discipline and traceability, procedural reflexivity, and user appropriation. For each criterion, I note which existing evaluation methods address it (for example, faithfulness metrics and factuality audits for evidence discipline, ambiguity benchmarks for ambiguity sensitivity, and calibration work for procedural reflexivity). I argue that these criteria could be reported and studied together under the heading of hermeneutic quality, rather than evaluating interpretive uses of LLMs solely in terms of local accuracy or plausibility.
    \end{itemize}
    \item 
    \textbf{With respect to societal dynamics:}
    \begin{itemize}[leftmargin=*]
        \item[(F)] 
        \textbf{Human-facing guidance for interpreting AI-mediated texts.}
        I translate this technical and conceptual proposal into human-facing guidance for what sometimes call digital hermeneutics: pedagogical and professional practices for interpreting, contesting, and appropriating AI-mediated texts. This includes treating model outputs as proposals rather than authorities, routinely requesting alternatives and explicit horizons, tracing sources, and using multiple perspectives or personas to surface disagreement. The aim is to keep human interpretive agency at the center of how people use AI to make sense of texts and situations, while still leveraging the model’s ability to retrieve, summarize, and recombine patterns across large corpora.
        \item[(G)] 
        \textbf{Domain-specific implications for law, education, and moral or cultural discourse.}
        I apply the notion of hermeneutic quality to three interpretive domains where LLMs are already in use: law and public policy (case summarization, statutory and constitutional interpretation); education and scholarly knowledge (explanations, literature reviews, textbook-like content); and moral, religious, and cultural discourse (scriptural exegesis, historical narratives, public moral debate). In each domain, I identify what typical LLM use currently misses—for example, the loss of dissenting legal opinions, the flattening of scholarly controversies, and the erasure of minority theological or historical perspectives. I then explain why these omissions matter socially, including for the legitimacy of legal reasoning, the cultivation of critical reasoning, and the protection of pluralism. Finally, I sketch how the proposed hermeneutic design patterns and quality criteria could reshape workflows and change how reasons and responsibilities are recorded.
    \end{itemize}
\end{itemize}

\subsection{Paper Organization and Scope}
 
The remainder of the article is organized as follows. Section~\ref{sec:hermeneutic-background} sketches modern hermeneutics as a theory of understanding as a historical, dialogical event and introduces the key figures and concepts used later in the paper. Section~\ref{sec:hermeneutics-meets-LLMs} reviews recent hermeneutic engagements with AI-generated text and LLMs and motivates the interaction-centered perspective adopted here. Section~\ref{sec:ai-mediated-loop} characterizes human--model interaction as an AI-mediated interpretive loop and distinguishes hermeneutic understanding from artificial (algorithmic) interpretation. Section~\ref{sec:trust_safety} develops the notion of a Hermeneutic Safety Loop and analyzes trust and safety risks and safeguards for co-interpretation. Section~\ref{sec:designing-hermeneutic-AI-systems} reorganizes existing LLM capabilities into design patterns for hermeneutically informed systems, proposes evaluation criteria for hermeneutic quality, and introduces digital hermeneutics as a user-facing literacy. Section~\ref{sec:societal-dynamics} applies this framework to law and public policy, education and scholarly knowledge, and moral, religious, and cultural discourse.

The paper does not attribute historical situatedness, intentionality, or hermeneutic understanding to LLMs. It proposes how to \textit{structure} human--AI (and more generally, multi-agent) interaction and evaluation so that hermeneutic desiderata, such as plurality, dialogue, historical situatedness, and the iterative character of interpretation, become operational and inspectable in practice, while interpretive responsibility remains with human agents.

\newpage

\section{Hermeneutic Background}
\label{sec:hermeneutic-background}

\vspace{-2pt}

\subsection{``Understanding'' as a Hermeneutical Event}

\vspace{-2pt}

Since its beginnings in biblical and classical exegesis, hermeneutics has wrestled with how we come to \textit{understand} a text, situation, or work of art. Early practices framed understanding as a methodological exercise: to recover an author’s intent or secure a fixed meaning. As hermeneutics evolved, however, philosophers came to treat interpretation as a historical and dialogical process. Interpreters never meet a text empty-handed; they bring prejudgments, cultural assumptions, and linguistic norms that condition what can be noticed, questioned, and learned. On this view, understanding is not the simple extraction of information. It is a \textit{transformative event}'', in which both the subject matter and the interpreter’s horizon are altered and brought into renewed focus. This dynamic interplay (usually called the hermeneutic circle'') shows how any \textit{part}'' becomes intelligible only against an anticipated \textit{whole}'', even as a revised sense of the \textit{whole} reshapes our reading of each \textit{part}. Understanding draws its depth from this historically conditioned, open-ended character: each new \textit{engagement} can unsettle inherited assumptions, widen insight, and provoke further questions. In this manner, ``understanding'' names an ongoing dialogue in which one’s perspective is continually tested, enriched, and reshaped by tradition, context, and the unforeseen nuances of the text or phenomenon at hand.

\vspace{-2pt}

\subsection{Key Figures and Concepts in Modern Hermeneutics}

Modern hermeneutics can be read as moving from a methodological program for interpreting texts to a philosophical account of how understanding is possible at all. The figures below mark a sequence of influential reorientations from interpretive \textit{technique} (Schleiermacher, Dilthey), to the \textit{ontology} of understanding (Heidegger, Gadamer), to a \textit{hermeneutics of the text and the self} (Ric{\oe}ur), and finally to a critique of closure and semantic presence (Derrida).

\vspace{-5pt}

\paragraph{Friedrich Schleiermacher (1768--1834).}
Schleiermacher is widely regarded as a father of modern hermeneutics \citep{osborne2011paradigm}, transforming it from a set of rules for interpreting specific texts (e.g., biblical exegesis) into a general hermeneutics of understanding linguistic communication.
He systematized hermeneutics as both an \textit{art} of understanding (emphasizing psychological intuition) and a \textit{methodological discipline} of textual interpretation (grounded in linguistic and historical analysis). 
His dual-axis framework distinguished between grammatical and psychological interpretation \citep{ricoeur1977schleiermacher, schleiermacher1998hermeneutics}:\vspace{-5pt}
\begin{itemize}[leftmargin=*]
    \item \textbf{Grammatical interpretation:} A systematic, ``comparative'' analysis of a text's linguistic usage, genre conventions, and historical-linguistic context.\vspace{-4pt}
    \item \textbf{Psychological interpretation:} A creative, ``divinatory'' act of empathetically reconstructing the author's subjectivity, intentions, and creative process.\vspace{-1pt}
\end{itemize}
By arguing that hermeneutics applies to \textit{all acts of communication}, he presented the discipline as one that extends beyond its traditional confines (biblical, legal, or classical texts).
At the center of his framework is the part--whole circularity of understanding (later discussed under the label \textit{hermeneutic circle}\footnote{%
The expression \textit{hermeneutische Cirkel} appears in the classical philologist August Boeckh's 1809 lectures on philological method; Schleiermacher later develops the part--whole circularity of understanding without adopting the expression ``hermeneutical circle.'' See \citep{boeckh1886enzyklopadie} and \citep{grondin2015hermeneutical}.}). This recursive interplay holds that understanding arises through an ongoing reciprocity between a text's \textit{parts} (e.g., words, sentences) and its \textit{whole} (e.g., genre, authorial intent, historical context).
In Schleiermacher, this circularity is complemented by a reciprocity between the interpreter's linguistic preconceptions and the attempt to reconstruct the author's historically situated intention and subjectivity. 
For Schleiermacher, this circularity is not a vicious circle to be avoided but the constitutive process through which understanding progressively unfolds.

\paragraph{Wilhelm Dilthey (1833--1911).}
Dilthey expanded hermeneutics into a foundational methodology for the human sciences (Geisteswissenschaften) \citep{dilthey1910aufbau, dilthey1972rise}, contrasting interpretive understanding (Verstehen) with the causal explanations (Erkl\"{a}ren) of the natural sciences (Naturwissenschaften).
For Dilthey, interpretation aims to reconstruct the \textit{inner life} (inneres Leben) or \textit{psychic nexus} (seelischer Zusammenhang) expressed in cultural productions, as well as the \textit{historically conditioned worldviews} (Weltanschauungen) sedimented in them. 
What makes such reconstruction possible is that lived experience (Erlebnis) becomes publicly accessible \textit{as meaning} through its \textit{objectifications} (Objektivierungen): texts, artworks, actions, and institutions.%
\footnote{Dilthey describes the ``objectification (of life)'' (Objektivierung) both as the externalization of human productivity (its becoming intersubjectively accessible) and as its embodiment in a shared ``sphere of commonality'' and ``universality''; for this latter dimension he borrows Hegel's term \textit{objektiver Geist} (``objective spirit'') \citep{dilthey2002formation}. See also \citep{tool2007wilhelm}.}
Accordingly, understanding (Verstehen) involves a reconstructive ``\textit{re-experiencing}'' (Nacherleben) of Erlebnis \textit{as objectified} in such expressions, and situating them within the historically shared medium that Dilthey, borrowing Hegel’s term, calls \textit{``objective spirit''} (objektiver Geist) \citep{dilthey2002formation}. 
By grounding interpretation in the relation between \textit{lived experience} and its \textit{objectification} within objective spirit, Dilthey presents understanding as historically situated: interpreters work within inherited forms of life rather than applying timeless laws to cultural meaning.

\paragraph{Martin Heidegger (1889--1976).}
Heidegger reframed hermeneutics by shifting its center of gravity from a methodology of textual interpretation to a fundamental-ontological inquiry into the question of Being (Seinsfrage). In \textit{Being and Time} \citep{heidegger1996beingtime}, he argues that \textit{understanding} (Verstehen) is not primarily a psychological or epistemological act but an \textit{existentiale}\footnote{Heidegger distinguishes between existential (ontological structures of Dasein) and existentiell (ontic or everyday aspects of human life).}---an existentiale of \textit{Dasein}---that is, a constitutive structure of its being-in-the-world (In-der-Welt-sein). Interpretation (Auslegung), by contrast, is the explicit ``working out'' of what is already understood more tacitly. Heidegger further argues that interpretation is guided by the \textit{fore-structure} (Vor-Struktur) of understanding: Vorhabe (\textit{fore-having}), Vorsicht (\textit{fore-sight}), and Vorgriff (\textit{fore-conception}) \citep{heidegger2008ontology}. This fore-structure names how understanding is always already oriented by prior involvement and inherited intelligibility; interpretation therefore does not begin from a neutral standpoint but unfolds from (and can revise) that prior orientation.
For Heidegger, the hermeneutic circle is not a methodological device but a structural feature of Dasein’s understanding: interpretation (Auslegung) explicates what is already understood (Verstehen) in terms of a fore-structure, and in explicating it can revise that understanding \citep{heidegger1996beingtime}.
By situating hermeneutics within the Seinsfrage, Heidegger presents Dasein as thrown (geworfen) into historically available meanings while projecting (entwerfend) possibilities of being; hermeneutics thus names the way Dasein’s being-in-the-world is disclosed and interpreted \citep{heidegger1996beingtime}.

\paragraph{Hans-Georg Gadamer (1900--2002).}
A student of Heidegger, Gadamer expanded ontological hermeneutics by foregrounding the \textit{historical} and \textit{dialogical} character of understanding in \textit{Truth and Method} \citep{gadamer2013truth}. 
In particular, he rehabilitates the concept of `\textit{prejudice}' (Vorurteil): historically formed fore-judgments are not simply distortions to be eliminated\footnote{Prejudice traditionally seen as a negative bias.} but conditions that \textit{enable} understanding. His notion of the ``\textit{fusion of horizons}'' (Horizontverschmelzung) names the way understanding occurs through an interplay between the interpreter's historically situated horizon and the horizon opened by the text within a living tradition, rather than by reconstructing an authorial standpoint as such. 
This process does not terminate in a definitive, context-free meaning; it remains an open-ended event sustained by a question-and-answer structure.
Crucial to his theory is \textit{effective-historical consciousness} (wirkungsgeschichtliches Bewu{\ss}tsein), which emphasizes that interpretive perspectives are inextricably informed by the history of transmission and prior interpretations \citep{gadamer2013truth}. Consequently, Gadamer portrays hermeneutics as a creative, ongoing conversation in which new insights continuously arise from the historically grounded encounter between interpreter and text.

\vspace{-5pt}

\paragraph{Paul Ric{\oe}ur (1913--2005).}
Ric{\oe}ur contends that phenomenology cannot remain at the level of immediate \textit{self-givenness} once meaning is mediated by symbols, discourse, and writing; it must therefore take a hermeneutic ``long route'' (long path) through the interpretation of expressions in which meaning is objectified \citep{ricoeur1975phenomenology, ricoeur1976interpretation}. Within this framework, he defines interpretation as a disciplined correlation of \textit{explanation} and \textit{understanding}: explanation reconstructs the internal organization of a text (e.g., by structural analysis), while understanding concerns what the text says by opening a referential ``world'' that can be inhabited and tested by the reader \citep{ricoeur1976interpretation, ricoeur1981hermeneutics, ricoeur1991fromtext}. Ric{\oe}ur formulates their relation through \textit{distanciation} and \textit{appropriation}. 
\textit{Distanciation} designates the effects of inscription by which discourse becomes fixed as a text and thereby acquires semantic autonomy with respect to speaker, original situation, and authorial intention; this autonomy is the condition under which the same text can be interpreted across new contexts \citep{ricoeur1976interpretation, ricoeur1981hermeneutics}. 
\textit{Appropriation}, the correlate of distanciation, names the reader's taking-up of the meaning and reference disclosed by the text—an act oriented to the text’s ``world'' rather than to psychological reconstruction of the author—through which self-understanding is mediated and potentially transformed \citep{ricoeur1991fromtext, ricoeur1992oneself}. 
Ric{\oe}ur extends this model beyond textual exegesis to the analysis of meaningful action and historical understanding: action can be ``read'' insofar as it is publicly intelligible and interpretable, and narrative provides a privileged form of configuration through which temporality and historical experience become describable and contestable \citep{ricoeur1981hermeneutics, ricoeur1991fromtext, ricoeur1984time}. 
In \textit{Time and Narrative}, he explicates this through the threefold structure of mimesis (\textit{prefiguration}, \textit{configuration}, \textit{refiguration}), where narrative emplotment (configuration) refigures the reader’s temporal experience and thereby supplies a framework for both fiction and historiography \citep{ricoeur1984time}. 
In \textit{Oneself as Another}, he connects interpretation to the constitution of personal identity by distinguishing \textit{idem}-identity (sameness) from \textit{ipse}-identity (selfhood) and by locating ``narrative identity'' at their intersection \citep{ricoeur1992oneself}. 
Finally, in \textit{Freud and Philosophy}, he contrasts a \textit{hermeneutics of suspicion} (unmasking distortion, rationalization, and ideology) with a \textit{hermeneutics of restoration} that seeks a renewed affirmation of meaning after critique \citep{ricoeur1970freud}.

\paragraph{Jacques Derrida (1930--2004).}
Derrida challenges hermeneutic approaches that presuppose a recoverable origin or a finally determinable meaning by reading texts for the conceptual hierarchies that organize them and for the points at which those hierarchies fail to secure themselves \citep{derrida1978writing, Derrida1976-DEROG, derrida2016grammatology}. 
He rejects treating \textit{deconstruction} as a transferable ``method'' applied from outside; it is a mode of reading that follows what a text must exclude, subordinate, or leave undecidable in order to sustain its own claims \citep{derrida1978writing}. 
A key term is \textit{diff\'{e}rance}---Derrida’s name for the structure of signification in which meaning arises through differences between signs and is deferred through chains of referral, so that no sign yields a self-identical presence of meaning \citep{derrida1973speech, derrida1978writing}. 
Accordingly, deconstruction attends to hierarchical oppositions (e.g., speech/writing, presence/absence) and shows that the privileged term depends upon what it excludes: ``presence'' is articulated through traces of non-presence, and ``speech'' is intelligible only through iterable marks that also characterize writing \citep{derrida1973speech, Derrida1976-DEROG, derrida2016grammatology}. 
Derrida describes this recurrent philosophical prioritization of presence as the ``metaphysics of presence'' and argues that it cannot be simply overcome by replacing one term with its opposite, because the opposition itself is structurally implicated in the text’s operation \citep{derrida1978writing, derrida2005end}. 
Read in this way, Derrida does not eliminate interpretation; he contests interpretive closure by showing that textual meaning is context-dependent and iterable, and therefore never exhaustively saturable by a single, final reading \citep{derrida1982margins}.

\vspace{-4pt}

\section{Hermeneutics Meets LLMs: Existing Scholarship}
\label{sec:hermeneutics-meets-LLMs}

\vspace{-2pt}

A small but growing body of work explicitly connects classical hermeneutics to LLMs, arguing that the tradition of interpretation offers tools for analyzing AI-generated text. Maciag \citep{maciag2024hermeneutics}, for example, contends that the ``research background of hermeneutics'' is indispensable for examining artificial texts produced by LLMs. On this view, the core hermeneutic questions of \textit{meaning}, \textit{context}, and \textit{understanding}, long applied to human-authored texts, can also be posed to machine-generated content. What follows sketches several hermeneutic perspectives on LLMs, focusing on Gadamer, Ric{\oe}ur, Derrida, and related work.

\subsection{Gadamerian Perspectives: Language, Tradition, and Dialogue}

Gadamer's philosophical hermeneutics, as developed in \textit{Truth and Method} \citep{gadamer2013truth}, treats understanding as an \textit{ontological event} rooted in \textbf{(i)} language, \textbf{(ii)} tradition, and \textbf{(iii)} dialogue. Building on this view, recent authors argue that LLMs do not genuinely ``understand'' language; instead, they simulate language use at the level of text. Pinell \citep{pinell2024does}, for example, asks whether generative language models ``have acquired natural language'' and, drawing on Gadamer, answers in the \textit{negative}. While LLMs can reliably produce text that reads as fluent, Pinell maintains that they ``lack the linguistically mediated reality that language provides.'' On this account, language is not exhausted by grammar; it is how humans \textit{engage} with the world and \textit{transmit tradition} \citep{pinell2024does}.

\vspace{1pt}

Pinell identifies four critical features of human language use that current LLMs lack: \textbf{(1)} groundedness in the world, \textbf{(2)} genuine understanding, \textbf{(3)} a community of discourse, and \textbf{(4)} a historical tradition \citep{pinell2024does}. 
Absent embodiment in a world, participation in a living community, and a cultural-historical horizon, an LLM’s outputs--however fluent---do not count as language in the full Gadamerian sense.
This reveals a hermeneutic gap between AI-generated text and the Gadamerian understanding of language: for Gadamer, language is not simply a formal structure, but a historically situated medium through which a community discloses a world. Gadamer's dictum, ``\textit{Being that can be understood is language},'' is often taken to imply that meaning emerges from a \textit{fusion of horizons}, the interplay between interpreter and tradition, which an isolated computational model cannot replicate on its own. 
Therefore, from a Gadamerian viewpoint, LLMs generate text but do not partake in the \textit{dialogical event of understanding}. 
Hence, their outputs should be treated as \textit{proposals for interpretation} to be questioned and revised, not as meanings that carry their own authority. This reminds us not to conflate statistical eloquence with true \textit{linguistic engagement} or \textit{conceptual thought}.

\subsection{Ric{\oe}urian Perspectives: Narrative and Interpretation}

Paul Ric{\oe}ur's hermeneutics emphasizes how narratives organize events into meaningful stories and how readers subsequently re-interpret these narratives. To date, direct academic engagements with Ric{\oe}ur's work in the context of LLMs remain scarce, recent commentators have invoked his ideas to elucidate AI-generated narratives. Central to these discussions is Ric{\oe}ur's \textit{threefold mimesis}, prefiguration, configuration, and refiguration, which can serve as a framework for understanding how LLMs generate and shape narratives \citep{ricoeur1984time}. For example, in a conceptual piece, Liggieri \citep{homoerectus2024} illustrates how \textit{raw data (prefiguration)} is processed by an AI model to \textit{configure} a textual narrative (e.g., composing a coherent story or answer), which is then \textit{refigured} by the human reader, who interprets and employs that output into lived context \citep{ricoeur1981hermeneutics, ricoeur1976interpretation}.

\vspace{1pt}

From this perspective, LLMs operate as ``narrative composers,'' creating story-like outputs, while the \textit{user} completes the hermeneutic circle by contextualizing and assigning meaning to these narratives. For instance, an LLM might generate a summary or advice (configuration) based on a given set of facts, yet it is ultimately the reader who determines whether that narrative is insightful, relevant, or ethical (refiguration) \citep{ricoeur1991fromtext}. On this Ric{\oe}urian view, \textit{meaning emerges through a process}, namely a collaboration between the AI's role in shaping a narrative and the human's role in interpreting it. Consequently, the evaluation of LLM outputs should address not only factual accuracy but also how effectively they assist users in \textit{making sense of information}, given that coherent narratives can influence individual and collective worldviews.
Moreover, Ric{\oe}ur's assertion that texts can be interpreted ``beyond the author's intention'' is particularly salient in the AI context, where no single human authorial intention (as reconstructed) governs the generated text. This suggests that AI texts call for a ``beyond-intention'' hermeneutic approach \citep{ricoeur1976interpretation, ricoeur1991fromtext}, in line with with Ric{\oe}ur's claim that a text's meaning can surpass what any author (or model designer) initially envisioned.

\subsection{Derridean and Poststructuralist Insights: Text, Difference, and Writing}
\vspace{-2pt}

Jacques Derrida's work, along with broader poststructuralist perspectives, develops a view of language in which \textit{meaning is never fixed but instead emerges from the interplay of differences and deferrals} (the ``play of signifiers''\footnote{%
    The ``play of signifiers'' is a pivotal concept in poststructuralist thought, especially in Derrida's work. It challenges the traditional Saussurean notion \citep{holdcroft1991saussure} of a stable signifier-signified relationship by arguing that meanings are produced through a continuous chain of signifiers that reference each other. This perpetual deferment of meaning, termed ``diff\'{e}rance,'' reveals the instability and fluidity inherent in language. For an in-depth discussion, see \citep{Derrida1976-DEROG}.}), 
famously encapsulated in Derrida's statement, ``\textit{there is no outside-text}.''\footnote{Note that ``il n'y a pas de hors-texte'' does not deny extra-textual reality. It denies unmediated presence, since reference is always mediated by textual and contextual traces.} This conception has been taken to parallel the way LLMs operate, and recent work has invoked Derrida's insights to elucidate how these models function. For instance, Vromen \citep{vromen2024language} proposes conceptualizing LLMs ``\textit{as semiotic machines rather than as imitations of human cognition},'' grounding his argument in Saussure's structural linguistics and Derrida's critique thereof. In this view, \textit{LLMs primarily model language as a system of signs\footnote{%
    Conceptualizing LLMs as ``semiotic machines'' highlights the idea that these models operate on the structural and relational aspects of language, focusing on the dynamics of signifiers rather than replicating the complexities of human thought. This perspective draws on both Saussurean linguistics and Derrida's critiques, positioning LLMs as tools that model language's intrinsic play and deferred meaning.}}, 
rather than replicating human thought (cognitive processes) \citep{vromen2024language}.

From this Derridean standpoint, LLMs encode what Derrida refers to as ``\textit{\`ecriture}'' (writing), an expansive network of references and traces, by statistically capturing patterns in text. Vromen argues the process of next-token prediction, which underlies LLM generation, ``\textit{effectively captures the dynamic nature of meaning}'' in language, insofar as each successive word is chosen relationally within an intertextual context rather than by appeal to a fixed presence of truth \citep{vromen2024language}. Consequently, LLM outputs exemplify the \textit{undecidability} and \textit{context-dependence} of meaning: they produce plausible linguistic continuations by leveraging learned differences rather than referring to an external, stable reality.
I take this claim in a deliberately limited, Derrida-consistent sense: what is ``captured'' is the \textit{play of signifiers} and the structured dependence of continuations on context, not \textit{hermeneutic understanding} or semantic reference secured by lived participation in a world. 
On my terminology, this supports treating outputs as \textit{interpretive proposals} whose significance remains underdetermined and must be appropriated, contested, and grounded by human interpreters within an explicit horizon.

Truman \citep{truman2024} describes transformer-based LLMs as ``digital hermeneuts,'' on the grounds that they compress vast corpora into webs of textual association. I do not adopt that label here. Instead, I reserve hermeneutic interpretation for humans and describe the model's contribution as artificial (algorithmic) interpretation: a pattern-based transformation of prompts and texts that offers proposals for human appropriation.
Inspired by Shannon's information theory \citep{shannon1948mathematical}, this researcher also emphasizes that \textit{information is not the same as meaning}. LLMs store extensive statistical data about language usage, but \textit{meaning emerges only through contextualized interpretation} (a point in line with Derrida's thinking).

In this view, Derrida-inspired accounts of textuality applied to LLMs frame these models as operating entirely within textuality and difference. This helps explain both their capability profile, since they encode a massive intertextual web, and their limitations, since they have no intrinsic anchor\footnote{externally provided referents/evidence constraints} to extra-textual reality or authorial intent. Extrinsic scaffolds (retrieval, sensors, ontologies) can impose such an anchor, but it remains designed rather than lived. The poststructuralist view thereby complements the Gadamerian one: where Gadamer stresses what LLMs \textit{lack} (world, tradition, dialogical understanding), Derrida-inspired researchers emphasize what LLMs \textit{are}, namely elaborate simulacra of language, whose significance remains \textit{unfixed} and depends on acts of interpretation.

\subsection{General Hermeneutic Approaches to AI Texts}

\vspace{10pt}

Beyond individual philosophical engagements, several researchers have adopted explicitly hermeneutic frameworks to interrogate AI-generated texts. Maciag's \textit{The Hermeneutics of Artificial Text} \citep{maciag2024hermeneutics} is an early effort to establish a ``hermeneutics of artificial text'' as a distinct research subfield. He argues that contemporary AI-generated texts often exhibit high local semantic fluency/plausibility, under distributional training, comparable to human prose, thereby warranting hermeneutic attention \citep{maciag2024hermeneutics}.
Maciag further asserts that centuries of hermeneutical scholarship, ranging from classical exegesis to the insights of Schleiermacher, Dilthey, and Gadamer, can clarify the interpretive dimensions of LLM outputs \citep{maciag2024hermeneutics}.

\vspace{4pt}

A central principle he highlights is the ``\textit{unique and autonomous character of each meaningful text, regardless of its origin},'' which takes on renewed significance for AI-generated texts \citep{maciag2024hermeneutics}. In other words, once a machine-generated text exists, it functions as any other text: it speaks through language, impinges on interpretive communities, and has the potential to influence readers. This autonomy justifies the application of established interpretive theories to AI outputs rather than dismissing them as derivative or as lacking interpretive significance.

\vspace{4pt}

At the same time, Maciag identifies specific interpretive complications. For instance, he argues that ``\textit{artificial text changes or even excludes the institution of context}'' upon which traditional texts rely \citep{maciag2024hermeneutics}. Human-produced discourse usually emerges within a broader cultural or intertextual network, while LLM outputs may appear ``orphaned,'' lacking the clear historical or cultural continuum that frames interpretation in human-authored works. This discontinuity, an interruption of the \textit{hermeneutic continuity}, calls for new strategies for interpreting AI content responsibly. Moreover, Maciag cautions that certain \textit{poetic functions} (e.g., semantic creativity or metaphorical richness) may be constrained by an LLM's reliance on patterned data rather than the open-ended semantic invention characteristic of human literary expression \citep{maciag2024hermeneutics}. Taken together, these insights show how \textit{AI-generated text both fits within and disrupts traditional hermeneutic categories}.

\vspace{4pt}

Other researchers have likewise turned to hermeneutic phenomenology to explore how humans engage with AI. One study adopts a \textit{hermeneutic phenomenological approach} to examine how workers incorporate LLM assistance into their professional routines, focusing on the lived experience and the interpretive processes that shape meaning-making in these contexts \citep{gonzalez2024beyond}. 
Here, the AI is not simply a text generator but an \textit{active agent}\footnote{I use ``agent'' here in a functional sense (an actant that shapes practices), not ascribing understanding, intentionality, or historicity to the system.} within human practices, with implications for our understanding, sense-making, and even identity in work environments. Delacroix \citep{delacroix2025designing, delacroix2025moral} further leverages hermeneutic concepts of \textit{sense-making} and \textit{meaning negotiation} to discuss the ways in which human moral discourse (conversations) might be altered by involving LLMs. 
She asks, ``\textit{What does it mean for human processes of reality construction and meaning negotiation to become intertwined with language models?}'' \citep{delacroix2025designing, delacroix2025moral}, pointing to a distinctly hermeneutic question about how our interpretive horizons intersect with machine-generated content. Collectively, these explorations show how hermeneutics supplies conceptual resources for analyzing AI-mediated texts in philosophical, social, and practical domains.

\section{AI-Mediated Interpretive Loop}
\label{sec:ai-mediated-loop}
\vspace{-2pt}

Human interaction with LLMs involves an \textit{AI-mediated interpretive loop}: 
the model performs an \textit{artificial (algorithmic) interpretation}\footnote{I reserve \textit{hermeneutic} for human understanding grounded in a lived historical horizon. I use \textit{artificial (algorithmic) interpretation} for a model's pattern-based mapping from prompts (and any retrieved context), conditioned by learned parameters from training, to generated tokens.} of the prompt and any retrieved context, conditioned by learned regularities from training, and the human performs the \textit{hermeneutic act} when reading, appropriating, and revising the output. The human interprets the AI's output, which itself was an \textit{algorithmic interpretation} (or transformation) of the human's input query and the data it was trained on. 
In effect, there are \textit{two coupled interpretation loops}: a model-side loop of functional interpretation and a human-side loop of hermeneutic interpretation. The model-side loop transforms the user's query (and any supplied or retrieved context) into a candidate continuation under learned regularities and system constraints; the human-side loop then situates, contests, and revises that output within a lived horizon. Across turns, the loops reciprocally condition one another: outputs reframe subsequent questions, and revised questions reframe subsequent generations. This feedback structure is resembles what social theory calls the \textit{double hermeneutic}: interpretive outputs feed back into subsequent practice (prompting, decisions, institutional justification), which in turn reshapes the next round of interpretation by influencing how users frame later prompts and how organizations authorize downstream action \citep{giddens1993new}.

\begin{figure}[t!]
    \centering
    \includegraphics[width=\linewidth]{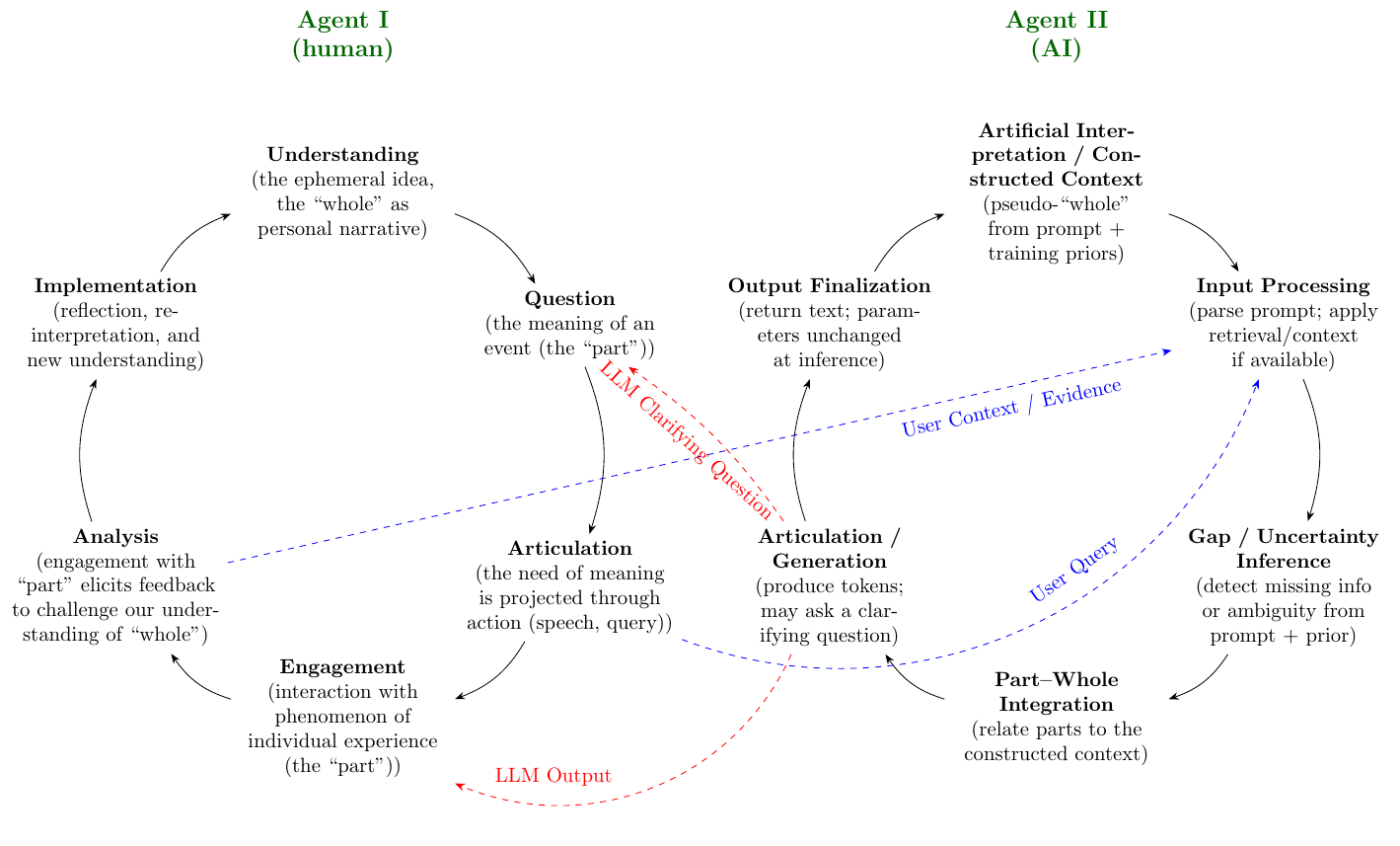}
    \vspace{-22pt}
    \caption{Two-agent interpretive loop (AI-mediated human hermeneutics). Left: the human hermeneutic cycle. Right: the model's artificial cycle. Solid arrows indicate clockwise, within-agent order. Dashed \textcolor{blue}{blue} arrows show Human$\rightarrow$AI hand-offs: the `User Query' from Articulation and optional `User Context/Evidence' from Analysis feed the model's Input Processing.
    Dashed \textcolor{red}{red} arrows show AI$\rightarrow$Human hand-offs: the `LLM Output' and optional `LLM Clarifying Question' (from Articulation/Generation) return to the human's Engagement and Question nodes, respectively.}
    \label{fig:two-agent-interpretive-loop}
\end{figure}

\vspace{-3pt}

\subsection{The Two-Agent Interpretive Loop}
\vspace{-2pt}

Figure~\ref{fig:two-agent-interpretive-loop} diagrams this two-agent interpretive loop. 
Agent~I (human) articulates a question and passes it, together with any contextual evidence from analysis, to Agent~II (AI), which performs an \textit{artificial interpretation}: it processes the prompt and any retrieved context, produces a candidate continuation conditioned on learned regularities, system constraints, and governance constraints, and (when prompted or configured) may emit a request for clarification before generating a reply.

\vspace{1pt}

Demichelis \citep[Sec.~3]{demichelis2024hermeneutic} argues that ``\textit{AI brings hermeneutics back into fashion}'' because of this very dynamic, suggesting we might speak of ``\textit{artificial interpretation}'' alongside artificial intelligence. For example, when an LLM is posed a question, it parses  the prompt, then generates an answer that user must read and interpret in turn. Misunderstandings can accumulate: if the model interprets the question out of context, it may give an irrelevant answer, which the user might misinterpret as well. \textit{Uncertainty} and \textit{ambiguity} are endemic in this loop. Delacroix points out that LLMs today often express unwarranted confidence, lacking the nuanced signals of uncertainty humans use in conversation \citep{delacroix2025designing, delacroix2025moral}. This can interrupt the hermeneutic process because if an AI states something without hedging, a user might not realize interpretation is still required and that the answer may be one of many possible meanings. The ethical implication is that AI systems should be \textit{designed to indicate uncertainty} or \textit{offer alternatives} (as some studies already do \citep{min2020ambigqa, kadavath2022language}), in order to keep the \textit{dialogue open} and invite the user's \textit{interpretive engagement} \citep{delacroix2025designing}. In moral or cross-cultural conversations, signaling uncertainty and clarifying the interpretive frame (e.g., assumed norms, audience, and scope) are crucial for genuine understanding, and if LLMs fail to do so, they risk ending the `conversation' prematurely and creating an illusion of finality.

\vspace{1pt}

In the human case, the horizon is lived and historically grown. On the model side, there is no such lived horizon, but there is still a technically enforceable boundary: prompts, retrieval corpora, ontologies, and normative constraints (e.g., policy rules, constitutional prompts) determine which questions are regarded as appropriate and which sources are treated as relevant. Section~\ref{sec:artificial-horizons} calls this boundary an \textit{artificial horizon} for the system and treats horizon design as a primary site of governance in LLM-mediated interpretation.

\vspace{-2pt}

\subsection{Illusion of Understanding and Epistemic Risks}

\vspace{-1pt}

A core concern is that LLMs can \textit{simulate} understanding without instantiating the conditions of understanding.  On Pinell's Gadamerian assessment, LLMs lack key conditions of understanding, such as groundedness in shared practices and participation in interpretive community/tradition \citep{pinell2024does}. Epistemologically, this means we should be cautious about trusting LLM-generated interpretations or analyses of texts. For instance, using a chatbot to interpret a poem or a legal document might yield a fluent explanation, but is the AI \textit{interpreting} or primarily \textit{recombining} learned textual regularities? The output may reflect pattern recombination rather than interpretive understanding.

Some authors argue that current AI lacks the \textit{understanding} or \textit{semantics} that characterizes human understanding. Floridi argues that such systems can manage texts statistically while understanding nothing of their content, decoupling apparent linguistic agency from intelligence \citep{floridi2023ai}. In other words, he described LLM outputs as \textit{zero-semantic}, that is, as syntax without semantics unless and until a human reader attributes meaning. This gap can lead users to over-ascribe comprehension to AI, a tendency Hofstadter discusses under the heading of the ELIZA effect \citep{hofstadter1995fluid} (people imbue the machine's words with more insight than is justified). Pinell's skepticism that LLMs can ever achieve natural language \citep{pinell2024does} suggests an ethical stance: we should not treat LLMs as autonomous meaning-makers on par with humans. In practical terms, it would be irresponsible, for example, to let an LLM's advice carry the same weight as a human expert's interpretation without human verification, especially in morally or culturally sensitive matters.
Floridi's point is used here as a caution against treating fluent generation as semantic competence: even when systems are externally grounded (e.g., by retrieval or ontological constraints), the relevant ``semantics'' for responsible practice remains mediated by human uptake, disciplinary norms, and institutional procedures for checking and attribution.
This is why this paper emphasizes artificial horizons and traceability: they do not grant the model understanding, but they make the conditions of use and the evidential status of its claims inspectable.

\vspace{-2pt}

\subsection{Authorial Intent and Accountability}

\vspace{-1pt}

Hermeneutics has long grappled with the role of authorial intention. With AI-generated text, however, the status of the ``\textit{author}'' is unclear: responsibility may be spread across developers, the algorithm, training corpora, and system design choices, rather than resting with a single intending agent. 
This ambiguity complicates both responsibility attributions and the interpretation of the text’s meaning. 
In intent-oriented strands of hermeneutics associated with Schleiermacher and Dilthey, a central interpretive question is what the author meant \citep{schleiermacher1998hermeneutics, dilthey1972rise}. 
Later hermeneutics complicates this picture by emphasizing historical mediation and the interpreter’s horizon \citep{gadamer2013truth}, while also attending to forms of meaning that exceed intention \citep{ricoeur1976interpretation}.
Henrickson and Mero{\~n}o-Pe{\~n}uela argue that computer-generated texts make clear that authorial intention is not sufficient for interpretation; such texts invite attention to meaning as it emerges in uptake and use, including forms of ``beyond-intention'' meaning \citep{henrickson2022hermeneutics}. 
This places a greater burden onto the reader (or user) to contextualize outputs responsibly.

Ethically, when an AI system produces a harmful or biased statement, it can be difficult to pinpoint accountability because the system is not \textit{intending} meaning in the human sense.
The interpretive process thus must include scrutiny of how the text may reflect biases in training data or design. 
This motivates transparency about data provenance (where available), retrieval sources, and system constraints, so readers can better situate and contest the output.

In legal interpretation, Coan and Surden caution against relying on LLM outputs as authoritative without verification and without attention to institutional accountability and justificatory standards \citep{coan2025artificial}. 
An AI might state interpretations with confidence but without verifiable evidential grounding, potentially misleading those who aren't critically evaluating its authority.

\subsection{Bias, Perspective, and the Need for Critique}

LLM outputs reflect the data they were trained on, which means they can carry forward biases present in society, including the reproduction of dominant or dominant cultural narratives \citep{bender2021dangers, gallegos2024bias}.  A hermeneutic (especially critical hermeneutic or \textit{hermeneutics of suspicion} \citep{ricoeur1970freud}) approach encourages us to interrogate AI outputs: \textit{Whose voice or perspective might this answer represent? What viewpoints might be excluded?} 
For example, if an LLM summarizes a historical event, how can we tell whether it is simply reproducing a Eurocentric narrative drawn from its sources? The model itself has \textit{no awareness} of perspective, which means it should not be relied upon to flag these biases without explicit prompting, tooling, or oversight; responsibility remains with the human interpreter.

Ethically, anyone using LLMs for research or teaching needs at least some training in how to interpret AI-provided output critically.
This is analogous to sourcing and contextualizing a quote from a book: we should ask, who wrote this and in what context? With AI, we must ask, what data and patterns produced this output? A number of authors have argued that transparency about sources and interpretive framing should be an ethical goal: AI systems could provide citations or explain their reasoning (structured rationales) in natural language (as some current systems already do \citep{singh2024rethinking}), allowing users to trace the origin of an answer, for example in citation-backed language models that support answers with verified quotes \citep{menick2022teaching}, and in philosophical accounts of technological mediation and the transparency/opacity of technical systems \citep{van2011between}.

\vspace{-2pt}

\subsection{Meaning and Truth in AI-Mediated Knowledge}
\vspace{-1pt}

Philosophically, once LLMs sit inside our epistemic practices, we have to ask: \textit{how do truth claims get established} in this new medium?
That is, LLMs change how we justify and present truth claims. In many public and professional practices, the warrant and legitimacy of \textit{truth-claims} are negotiated through \textit{dialogue}, \textit{evidence}, and institutional forms of \textit{consensus over time} \citep{mccarthy1991theory, o2005telling}, all of which are interpretive processes. LLMs, by contrast, produce answers by drawing on statistical associations, which can make the \textit{appearance of meaningfulness} difficult to tell apart from well-supported claims. 
A hermeneutic outlook treats interpretation as always provisional and situated. On this view, take LLM outputs as \textit{proposals}---as parts of an ongoing conversation---rather than as final truths.
This is especially important in interpretive domains: e.g., if an AI gives a certain interpretation of a literary character or a religious passage, it should ideally be a starting point for discussion, not the end. Gadamer spoke of the ``fusion of horizons'' \citep{gadamer2013truth}, meaning that understanding happens when the horizon of the text and the horizon of the reader come together. A related question is whether users risk aligning their horizon uncritically with the model's data- and prompt-bounded horizon. If users accept AI outputs at face value, the rich hermeneutic process of back-and-forth (questioning, contextualizing, revising interpretations) could be shortchanged. 
Yet \textit{constructive interaction} remains possible: an AI may surface interpretations or information we had not considered, expanding our horizon—provided we engage with and integrate them thoughtfully rather than accept them passively.

LLMs can be useful in interpretive work, but their use comes with \textit{responsibility}. These models do not \textit{truly know} in the human sense; they generate text that is plausible. For that reason, the ethical imperative is to \textit{treat AI-generated interpretations as fallible, partial, and in need of human contextualization}. Hermeneutics--long attentive to the reader/interpreter's active role---offers a lens for users to engage LLM outputs critically rather than taking them at face value.

\section{Trust and Safety in the Co-Interpretive Paradigm}
\label{sec:trust_safety}

This section argues that safety-relevant progress in LLM-mediated interpretation cannot be reduced to increased autonomy or improved performance on standard task metrics such as accuracy and preference win rates. In interpretive domains, safety also depends on whether human interaction with the system makes interpretive claims, assumptions, and value trade-offs explicit enough to be contested, audited, and attributed to identifiable decision-makers. This reframes how trust and safety should be understood. In paradigms that treat the system as if it were autonomous optimizer, safety is treated primarily as an external constraint on a system whose primary objective is to optimize a performance metric. In a co-interpretive paradigm, safety is treated instead as a property of the interpretive procedure that governs model updates and downstream action. The system justifies trust not by asserting correctness, but by participating in a logged, reviewable sequence of steps in which interpretations are surfaced, compared with alternatives, revised when challenged, and finally authorized by accountable humans.

\subsection{The Hermeneutic Safety Loop}

I operationalize this stance via the \textit{Hermeneutic Safety Loop} (HSL), a workflow that structures how humans and AI refine goals, interpretations, and interventions before any high-impact or hard-to-reverse commitment\footnote{For example, decisions with legal, clinical, or institutional consequences; model updates; policy actions; deployments affecting access/eligibility.} is made. In HSL, the basic unit is the logged interpretive episode---not the individual model output---documenting specified inputs, intermediate outputs, and the final decision.

\vspace{1pt}

Given a task context and a candidate action such as a policy recommendation, a model update, or a deployment change, an HSL episode has five steps. First, human participants and the system elicit multiple plausible interpretations of the context, objectives, and constraints, so that the interaction produces an explicit set of competing framings rather than a single privileged reading. Second, for each interpretation they record a structured ledger of normative commitments and epistemic assumptions, including fairness targets, acceptable harm thresholds, privacy boundaries, assumptions about data representativeness and domain applicability, and relevant institutional or legal constraints. Third, the system produces a contrastive justification document that compares interpretations (and, where appropriate, declines to rank them when the preference depends on contested values), linking each to supporting evidence (citations/quotations) when available. Fourth, humans and system select an operative interpretation in a way that preserves residual disagreement and uncertainty rather than smoothing it away; contested framings, thresholds, and open questions are treated as safety-relevant information and are carried forward in the record of the episode. Fifth, any resulting commitment to a training change, deployment decision, or governance action is tied to a traceable chain of interpretive steps and to a clearly identified human decision-maker, and the episode log records who authorized which choice, on the basis of which interpretation and assumptions, together with monitoring targets and rollback conditions where appropriate.

\vspace{1pt}

The HSL shifts safety analysis from static properties of a model to the traceable process of meaning-making that justifies how the model is updated and used. It turns interpretive work, which is often treated as informal domain expertise around a system, into a first-class workflow that can be inspected and audited.

\subsection{Hermeneutic Pathologies of Co-Interpretation}
\label{subsec:hermeneutic-pathologies}

The Hermeneutic Safety Loop treats safety as a property of an interpretive procedure rather than as a static characteristic of a model. That procedure, however, can itself go wrong in structured ways. In this subsection I gather a set of recurrent failure modes into what I will call \textit{hermeneutic pathologies of co-interpretation}.
By \textit{hermeneutic pathology} I mean a systematic breakdown in one or more of the hermeneutic conditions emphasized earlier---effective-historical consciousness, horizon plurality, part--whole integrity, and appropriation---when humans and AI systems co-produce interpretations. These failures are not simply local mistakes or adverse outcomes in the usual machine-learning sense; they are distortions in how meaning is formed, situated, and attributed within the human–AI interpretive loop.

There is a family of characteristic pathologies of human--AI co-interpretation that is most perspicuously characterized in hermeneutic terms, because they concern breakdowns in horizon plurality, part--whole integrity, and appropriation rather than only errors or harms in the usual ML sense.
In what follows I describe six such pathologies and indicate which hermeneutic conditions they primarily violate.

\vspace{-2pt}

\paragraph{Overreliance drift: loss of effective-historical consciousness.}
Overreliance drift occurs when repeated exposure to fluent and apparently well-reasoned outputs leads users to reduce scrutiny and to accept model interpretations with progressively less checking. Human oversight remains nominal but loses substance. Hermeneutically, this is a loss of \textit{effective-historical consciousness} in Gadamer's sense \citep{gadamer2013truth}: users cease to experience their own prejudgments and horizons as operative in the interpretive event and instead treat the system's outputs as if they lacked a situated standpoint. Appropriation in Ric{\oe}ur's sense \citep{ricoeur1976interpretation} is weakened: instead of integrating and critically owning an interpretation, users defer to the model's proposal as a ready-made meaning. The risk is an \textit{illusion of understanding} in which both human and system appear to understand while hermeneutic responsibility has effectively been surrendered.

\vspace{-2pt}

\paragraph{Authority laundering: systematic underrepresentation of authorship and responsibility.}
Authority laundering arises when AI-generated outputs are treated as neutral expert judgment even though they encode specific value choices, corpus biases, and modeling assumptions that have not been surfaced or debated. Hermeneutically, this is a pathology of \textit{authorship and imputability}. In Ric{\oe}ur's terms, texts acquire autonomy through distanciation, but their appropriation still calls for assigning responsibility and locating a speaking position \citep{ricoeur1976interpretation,ricoeur1991fromtext}. In authority laundering, responsibility is dispersed across training data, model design, and prompting conventions in such a way that no interpretable ``author'' remains to whom a meaning can be imputed. Organizational actors then invoke the system as a source of authority without acknowledging their own role in configuring its pseudo-horizon. Interpretive responsibility is laundered through the system.

\vspace{-2pt}

\paragraph{Interpretive collapse: suppression of plurality and \textit{diff\'erance}.}
Interpretive collapse appears when human and AI participants converge too quickly on a single framing of a problem without systematically exploring alternatives. The interaction produces one apparently coherent reading while suppressing others. From a Gadamerian perspective, this is a breakdown of \textit{horizon plurality}: the fusion of horizons becomes a premature closure, not a dialogical negotiation across perspectives. From a Derridean perspective, it is a suppression of \textit{diff\'erance}, understood as the play of differences and deferrals that keeps meaning open \citep{derrida1978writing}. The pathology consists in treating one possible construction as the natural or inevitable meaning, thereby erasing the plurality of readings that hermeneutic practice seeks to preserve.

\vspace{-2pt}

\paragraph{Benchmark overfitting: replacing hermeneutic quality by narrow metrics.}
Benchmark overfitting occurs when the quality of co-interpretation is evaluated primarily in terms of what is easy to measure, such as local accuracy, latency, click-through rates, or user satisfaction scores, while hermeneutic virtues such as part--whole integrity, ambiguity sensitivity, and evidential support and traceability are neglected in practice. In Dilthey's terms, explanation (Erkl\"aren) displaces understanding (Verstehen) as the \textit{dominant norm} in domains where understanding should retain priority \citep{dilthey1972rise}. Metrics that capture narrow predictive performance are allowed to stand in for the richer notion of \textit{hermeneutic quality} introduced earlier. The pathology is not that metrics are used, but that they become the de facto horizon within which interpretive success is defined, crowding out qualitative and historically informed judgments.

\paragraph{Value lock-in: ossifying one artificial horizon as if it were tradition.}
Value lock-in arises when early decisions about objectives, constraints, and acceptable trade-offs become path-dependent as systems and institutions scale, making revision difficult even when new evidence or social concerns would warrant it. In terms of artificial horizons, a particular configuration of sources, scopes, and value commitments is gradually treated as if it were a natural continuation of tradition rather than a contestable technical document. Gadamer emphasizes that traditions remain open to reinterpretation and that prejudices can and must be revised in dialogue \citep{gadamer2013truth}. Value lock-in turns an artificial horizon into a pseudo-tradition that resists critique, undermining both horizon plurality and the possibility of future fusion of horizons. The system continues to ``speak'' from one frozen background even as the human lifeworld moves on.

\paragraph{False transparency: pseudo-explanations that close conversation.}
False transparency occurs when systems present high-volume explanations or visualizations (e.g., long rationales, attribution displays, dashboards) that signal openness yet do not increase the interpreter’s ability to audit the operative assumptions, trace evidential provenance, or contest the decision. Here, explanatory volume substitutes for meaningful contestability. Hermeneutically, this is a pathology of dialogue: rather than sustaining the question–answer openness that Gadamer takes to be constitutive of understanding \citep{gadamer2013truth}, the system offers the appearance of dialogue while foreclosing the routes by which reasons can be challenged, revised, or rejected. The user is given rationales, but not warrant-bearing reasons of the kind needed for critique. Ricœur’s appropriation is again at stake: the interpreter cannot make the explanation their own because the points at which it could be tested—assumptions, evidence links, and decision thresholds—remain inaccessible. The exchange thus takes the form of interpretation without enabling its critical uptake.

\paragraph{From pathology to design.}
These pathologies show that the mere \textit{presence of humans in the loop does not by itself ensure safety}. Co-interpretation can fail hermeneutically even when local error rates remain low. The Hermeneutic Safety Loop and the design principles in the next subsection can be read, in part, as efforts to prevent or mitigate these failures: explicit artificial horizons and horizon design guard against value lock-in and authority laundering; mechanisms that surface alternative horizons and interpretations counter interpretive collapse; metrics for hermeneutic quality resist benchmark overfitting; and structured contestability with traceable interpretation addresses overreliance drift and false transparency by keeping human appropriation and responsibility visible and auditable.

\subsection{Design Principles for Trustworthy Co-Interpretation}

The risk of harm from AI is, to a significant extent, the risk that computer systems produce or perpetuate harmful outcomes without the moral and ethical judgment that humans typically apply in decision-making. I argue that trust, safety, and responsibility should be treated as design and governance requirements rather than as marketing ``value adds,'' and that contestability by design \citep{alfrink2023contestable} should be treated as a primary principle for ensuring that systems making consequential decisions remain open to challenge. By ``\textit{contestability by design}'', I mean that a system is engineered so that it is procedurally straightforward to pose counter-questions, offer alternative interpretations of the same data, and record explicit conditions under which a decision would be revisited. Thus, if a user queries a recommendation provided by an ``intelligent'' search engine, the system should support a structured challenge-and-response process that either revises the recommendation in light of the challenge or states what additional evidence or arguments would be required to justify a subsequent change.

\vspace{1pt}

Traceable interpretation means that the evolution of \textit{goals}, \textit{assumptions}, \textit{evidence}, and \textit{decisions} throughout an HSL episode should be logged so that later audits can reliably reconstruct what occurred. In particular, the record should make clear which interpretations were considered, which assumptions and value commitments they recorded, what justifications and evidence supported the final choice, and who authorized that choice. Disagreement preservation is the requirement that unresolved divergence---between human participants, between a human participant and the AI system (e.g., a conversational bot), or between different personas of the same system---be recorded as part of the output and retained in the record, rather than erased during aggregation. This way, the users involved, as well as downstream applications and oversight bodies, can see where disagreement occurred, which positions or conclusions were at stake, and how that disagreement was handled.

\vspace{1pt}

Human-agency constraints restrict which decisions the system may make without human authorization. In particular, changes to high-level objectives and deployment decisions in safety-critical settings should require human review and explicit sign-off on the resulting safety trade-offs. System recommendations and analyses may inform these decisions, but they function as proposals: final approval must come from pre-specified levels of organizational responsibility. These approval levels should be explicitly defined at least for (i) changes to the system’s high-level goals and (ii) deployment in safety-critical settings.

Calibration matters for large language models because fluent output can be wrong, especially in under-determined cases where multiple interpretations are consistent with the available evidence. We therefore adopt a principle of calibration over persuasion: training objectives and evaluation metrics should reward the model for aligning its communicated confidence and claims with what the evidence supports. Concretely, the model should (i) calibrate confidence to demonstrated reliability, expressing lower confidence in settings where it has empirically higher error rates on similar inputs and higher confidence where it has empirically lower error rates; and (ii) calibrate confidence to under-determination, reserving high-confidence claims for cases where the evidence strongly supports a conclusion, and otherwise qualifying conclusions and presenting plausible alternatives. The success criterion is not persuasive force but whether stated confidence and the handling of ambiguity track empirical error rates and reflect the structure of the underlying interpretive problem.

\vspace{1pt}

These principles give the Hermeneutic Safety Loop a concrete shape in system and process design. They embed interpretation sets, assumption and value ledgers, contrastive and sensitivity-based justifications, disagreement records, and accountability mechanisms into the way co-interpretive systems are built and operated.

\subsection{Implications}

This framing changes the role that explanations play in human interaction with LLMs. Explanations are not optional add-ons around a fixed system; they instantiate the interpretive procedure through which use and improvement become corrigible and socially accountable. The relevant question is therefore not simply whether a model can generate an explanation, but whether the explanations and associated logs support a Hermeneutic Safety Loop episode that is auditable, contestable, and grounded in identifiable human agency.

Because of the nature of co-interpretive systems, evaluating them cannot be based on counting the number of explanations produced, nor on the linguistic quality of these explanations. Evaluation should ask whether it is possible to \textit{reconstruct} the interpretive steps and value choices taken in the process from logs, whether the user can effectively challenge and revise the recommendations provided by the system, and whether critical decisions remain under user control. Finally, evaluating system safety must be based not only on performance indicators, but also on the quality and integrity of the interpretive procedures used to produce the results.

\section{Designing Hermeneutic AI Systems}
\label{sec:designing-hermeneutic-AI-systems}

I draw on the opportunities and challenges I have identified so far to propose ways to integrate \textit{hermeneutic principles} into the design, evaluation, and use of LLMs. Although research at this intersection remains limited, the design patterns and evaluation criteria that follow---some established and others exploratory---indicate how a \textit{hermeneutically informed} approach can be made inspectable and enforceable in LLM-mediated workflows.

Before turning to concrete heuristics, it is worth noting that contemporary LLM interfaces already exhibit partial implementations of what I describe here. Users can ask for summaries from a particular perspective, request multiple interpretations of a passage, refine answers in multi-turn dialogue, or provide domain-specific context via retrieval. I do not claim novelty for these capacities. The contribution is to reorganize them under a hermeneutic perspective, making explicit which patterns foster plurality, historicity, and part–whole integrity, and which encourage an illusion of understanding.

\subsection{Dialogical and Participatory Interfaces}

A core Gadamerian claim is that understanding emerges through dialogical, back-and-forth engagement rather than unidirectional information transfer. In Gadamer's terms, meaning arises in the ``fusion of horizons'' through a continual interplay between interlocutors' perspectives \citep{gadamer2013truth}. Contemporary LLMs often default to a simple question--answer dynamic. To encourage genuinely \textit{interpretive} engagement, systems must be designed to facilitate a more \textit{dialogical}\footnote{Here `dialogical' names the interaction form (question–answer–counterquestion), while `participatory' names interface affordances for annotating, revising, and contesting outputs.} mode of interaction.

\paragraph{Participatory Interfaces.}%
In line with Delacroix's suggestion for participatory interfaces \citep{delacroix2025designing, delacroix2025moral}, we can build platforms where users annotate AI-generated outputs in real time, highlighting unclear or ambiguous parts and requesting further clarifications\footnote{%
    For example, a feature of an application may \textit{enable users} to mark parts of the AI's output as unclear and/or ask the model to clarify its assumptions. This would mimic \textit{hermeneutic dialogue} by requiring the model to revisit, adjust, and refine its response in light of user feedback, much as interlocutors in dialogue revise their contributions in pursuit of mutual understanding.}. 
Thus, the interface would mimic the historically informed dialogue that hermeneutics values. Users could click to flag passages of an AI-generated summary as ``misaligned with their prior knowledge'' or as ``lacking cultural context''. The system would then revise or expand the summary in response, sustaining a cycle of clarification and revision through which meaning is \textit{co-constituted}\footnote{%
    That is the output is iteratively revised in response to user contestation and added evidence.%
} in a loop of clarification and revision\footnote{%
    By co-constituting interpretive moves between humans and AI \citep{delacroix2025designing, delacroix2025moral}, such systems turn the use of LLMs into a collaborative interpretive process. In practice, this might look like AI-assisted analysis tools where the AI offers several interpretations or options, and the human guides which is most meaningful. The goal is a system that does not simply return an answer, but structures interaction so that the user must interpret and interrogate the answer, emulating the back-and-forth of human conversation that Gadamer sees as central to understanding.}.

\paragraph{Iterated Hermeneutic Reflection.}
Rather than having a fixed output after a single pass, an LLM can be used to provide multiple rounds of reinterpretation in what we call a \textit{multi-round interpretive flow}. We might ask the LLM to provide a second or third reading of the same material, with each new reading incorporating user feedback. This \textit{multi-round interpretive flow} models the reciprocal nature of hermeneutic inquiry, with each revision serving as an occasion to test, refine, or expand the partial understanding reached so far. Such designs acknowledge that meaning is rarely final and is instead revised through iterative engagement.

\paragraph{Auto-Commentary and Self-Questioning:}
Chain-of-thought (COT) prompting \citep{wei2022chain, lyu2023faithful, zhang2022automatic} and self-explanation features \citep{huang2023can} have received growing attention in recent work. Drawing on medieval philosophical commentary, we propose to instruct an LLM to generate a ``commentary'' on its own output. Concretely, the commentary would (i) flag potential ambiguities, (ii) state the assumptions required for its reading of the input, and (iii) link key claims to cited spans in the input. This remains far from the kind of self-reflection that Gadamer or Ric{\oe}ur treat as central to hermeneutic interpretation. Even so, such auto-commentary can simulate a more critical posture toward the model’s output and may encourage users to engage it more actively and skeptically, without conferring epistemic authority on the system.

\subsection{Artificial Horizons, Ontologies, and Horizon Design}
\label{sec:artificial-horizons}

Hermeneutic traditions understand a \textit{horizon} as the background of expectations, prejudgments, and lived history that makes understanding possible. Gadamer's notion of \textit{effective-historical consciousness} emphasizes that every act of understanding is framed by one's cultural and historical situation \citep{gadamer2013truth}. For a human interpreter, this horizon is neither fully explicit nor fully specifiable; it belongs to a historically situated life.

LLMs, by contrast, generate content without a lived horizon or the dialogical engagement that Gadamer treats as essential for understanding. They lack an \textit{intrinsic} horizon and the kind of participation in tradition that hermeneutics regards as constitutive of understanding. For an LLM-based system there is no such lived horizon, but there is still a determinate technical boundary: the range of sources, assumptions, and norms that are actually enforced when the system is used in a given domain. I call this boundary an \textit{artificial horizon}. What we can provide instead of genuine historicity is an extrinsically imposed horizon, for example via context headers, ontologies, curated corpora, or value statements that shape generation. Although this does not endow the model with effective-historical consciousness, it leverages the hermeneutical insight that \textbf{context shapes meaning} and makes that context visible, configurable, and contestable.

\paragraph{Definition.}
By an \textit{artificial horizon} I mean an explicit, versioned specification that satisfies four conditions:
\begin{itemize}[leftmargin=2em]
    \item[(i)] 
    \textbf{Background commitments.} 
    It specifies the \textit{background assumptions}, \textit{scope constraints}, and \textit{normative commitments} that apply within a given domain. Examples include the relevant jurisdiction and court level in legal applications, the temporal coverage of the corpus, the doctrinal or theoretical stance treated as default, and the institutional policy goals or values the system is expected to respect.
    \item[(ii)] 
    \textbf{System conditioning.} 
    It is used to condition model calls within that domain configuration. Prompts, retrieval filters, system messages, and value overlays are constructed so that generation is constrained by this specification rather than by an unspecified global default.
    \item[(iii)] 
    \textbf{User representability.} It is presented in a form that intended users (e.g., students, caseworkers, attorneys) can inspect. This may be a visible header, a configuration panel, a profile description, or documentation that is directly linked to the interaction. Users do not have to infer the operative horizon from indirect cues.
    \item[(iv)] 
    \textbf{Revisability and traceability.} It can be changed only through a defined procedure, and changes are logged. It is always possible to see which version of the horizon was in force for a given interaction, who altered it, and when.
\end{itemize}

The horizon is ``artificial'' because it is engineered as a technical artefact rather than lived as a historical horizon. It nevertheless determines which kinds of questions the system treats as well formed, which sources it is allowed to draw on, and which families of answers are treated as appropriate in that domain.

\paragraph{Horizon Engineering.}
I use the term \textit{horizon engineering} to refer to the design and maintenance of artificial horizons. It involves at least three design decisions:\vspace{-2pt}
\begin{itemize}[leftmargin=*]
    \item 
    First, one must determine which commitments the horizon specification should include—for example, which legal sources are treated as authoritative; which doctrinal family or theory a theological or philosophical assistant presupposes; which ontologies or curated corpora delineate the domain; and which policy constraints and institutional values govern a deployed system.\vspace{-2pt}
    \item 
    Second, one must decide how this specification is enforced in practice, for example through fixed context headers, retrieval filters, fine-tuning, or value overlays.\vspace{-2pt}
    \item 
    Third, one must define who has the authority to change the horizon, how those changes are documented, and how disagreements about the horizon are handled institutionally.
\end{itemize}

In this sense, horizon engineering is a primary site of design and governance for LLM-mediated interpretation rather than a purely technical detail of prompt engineering.

\vspace{-2pt}

\paragraph{Ontologies as Pseudo-Traditions.}
Echoing Schleiermacher’s thesis that every act of interpretation is constrained by some pre-understanding (Schleiermacher, 1998), one straightforward way to construct an artificial horizon is to pair a domain ontology with a curated corpus. The ontology stabilizes the core concepts, relations, and categories the system is meant to use, while the corpus determines which texts count as authoritative or representative. For example, a legal assistant might be configured with (i) an ontology of legal concepts for a given jurisdiction, (ii) a closed list of statutes, regulations, and case reports, and (iii) an instruction to interpret questions according to a specified approach to legal interpretation.
This configuration is recorded and shown at the top of the interface. Whenever the model answers, it does so only by citing or summarizing from the allowed corpus and by using the terms defined in the ontology.\footnote{%
    An organization may want an AI system to draft guidance that reflects its mission and values. The prompt can embed an ontology that defines the organization’s key terms and codifies those values.}

In implementation terms, this corresponds to a combination of a system prompt, retrieval filters, and fine-tuned behaviour \citep{cai2018skeleton, lewis2020retrieval}. The point of calling it an artificial horizon is to treat that configuration as a first-class object of design and governance. Changing the horizon is then a normative decision about which sources, scope limits, and interpretive stance are appropriate, not a purely technical tweak.

\paragraph{Cultural and Ethical Overlays.}
Beyond domain-specific ontologies, institutions and communities often hold ethical principles and cultural norms that condition how meaning is formed and negotiated. Drawing on Dilthey's emphasis on historically situated worldviews \citep{dilthey1972rise}, we can equip prompts with cultural overlays or explicit guidelines that reflect an institution's identity. A healthcare organization may, for example, supply a patient-centred charter to guide clinical decision-support dialogues. A theological community might provide key doctrinal tenets before soliciting AI-generated exegesis. These overlays connect machine-generated text to the community's shared background, and they help ensure that outputs reflect the \textit{prejudices}, in Gadamer's rehabilitated sense of the term \citep{gadamer2013truth}, that inform real-world interpretive practice.

On an artificial-horizon view, these overlays are not optional add-ons. They function as background commitments: built into the specification, enforced through prompts and retrieval, and made visible in the interface. For that reason, they are open to revision and dispute on the same footing as decisions about sources and scope.

\paragraph{Hermeneutic Adequacy and Corruption.}
The same apparatus can be used well or badly. From a hermeneutic perspective, an artificial horizon is \textit{hermeneutically adequate} when at least three conditions are met:
\begin{itemize}[leftmargin=1.8em]
    \item[(1)] \textbf{Explicit selectivity.} The selectivity of the horizon is made visible. Users can see which jurisdiction, doctrinal family, corpus, or value commitments are in force and which plausible alternatives are not included.\vspace{-2pt}
    \item[(2)] \textbf{Support for alternatives.} The system makes it practically possible to invoke and compare alternative horizons for the same task. A hermeneutically adequate system should make it practically possible to invoke, for example, switch between different legal schools, theological traditions, or historiographical frames, and see how the answers change.\vspace{-2pt}
    \item[(3)] \textbf{Contestability.} Users can in fact contest the operative horizon. They can request a different horizon, propose changes, or escalate disagreements about the horizon to identifiable human decision-makers who are responsible for accepting or rejecting those changes.
\end{itemize}

\noindent
By contrast, an artificial horizon is hermeneutically corrupting when one or more of the following holds:
\begin{itemize}[leftmargin=1.8em]
    \item[(a)] The horizon that actually conditions generation is hidden or only indirectly inferable. Users see a neutral-seeming interface while strong constraints operate behind the scenes.\vspace{-2pt}
    \item[(b)] The design makes it difficult or impossible to access alternative horizons for the same domain, even when the domain is in fact plural (for example, where multiple doctrinal or methodological traditions coexist).\vspace{-2pt}
    \item[(c)] The interface presents the operative horizon as if it were natural, neutral, or inevitable, rather than as one contestable specification among others.
\end{itemize}

For example, in a legal assistant, the artificial horizon might be defined by parameters that restrict the system to United States federal appellate cases after a certain date, that direct statutory interpretation through a predominantly textualist approach, and that treat specified canons and treatises as authoritative. In a theological assistant, the artificial horizon might be defined by a particular confessional tradition, a canon of magisterial sources bounded by that tradition, and a doctrinal ontology appropriate to it. Horizon engineering, in both cases, concerns how these horizon-defining choices are specified and made visible, how they are enforced in prompts and in search and retrieval (including ranking and filtering), and how users can move to an alternative horizon—such as another jurisdiction or another confessional tradition—when the interpretive task requires it.

From a Gadamerian perspective, an artificial horizon does not reproduce the effective-historical consciousness of a tradition or lived participation in it. However, an artificial horizon fixes the formal structure that a horizon has within hermeneutics: it determines the framework of questions and possible answers, and the background commitments that have to be operative for the system to ``speak''. This is the level at which we can approach horizon engineering as a safety-critical systems engineering activity for LLM-based interpretation systems.

\vspace{-4pt}

\subsection{Encouraging Plurality: Contrastive and Perspectival Outputs}

\vspace{-2pt}

A key insight from hermeneutics, and particularly from Ric{\oe}ur's distinction between a \textit{hermeneutics of suspicion} and a \textit{hermeneutics of faith} \citep{ricoeur1970freud}, is that interpretive richness often emerges when multiple and sometimes conflicting readings are held in productive tension. LLM-based tools can be deliberately designed to elicit \textit{plural} interpretations and thereby foreground the open-ended character of meaning that both Gadamer and Derrida emphasize.

\vspace{-7pt}

\paragraph{Contrastive Interpretations.} 
Rather than returning a single answer, a system may output multiple contrasting interpretations of the same input. One example is the Shakespearean Sonnet reader in the accompanying panel. It produces three readings: a romantic reading that emphasizes love and beauty; a political reading that foregrounds power relations and gender roles; and a Derridean reading that questions the stability of any fixed meaning.
By staging a \textit{dialogue of interpretations}, the interface presents users with contrasting views to compare and critically evaluate. This can be read as analogous to Gadamer's fusion of horizons, in which understanding emerges through the encounter of different horizons, and it reflects Derrida’s insistence on the \textit{undecidability} of meaning, that is, the claim that meanings remain open rather than finally fixed.

\vspace{-7pt}

\paragraph{Perspectival Prompts.}
Technically, LLMs can shift their interpretive frame by adopting different ``personas'' or ``lenses'' when prompted \citep{wang2023rolellm}. A user can ask the model to answer as a Marxist critic, then as a psychoanalytic theorist, then as a legal historian. This deliberately \textit{multivocal} pattern makes explicit that all readings, including AI-generated ones, are shaped by specific pre-understandings and fore-structures of interpretation in the Heideggerian sense. Instead of collapsing perspectives into a single authoritative account, the interface allows users to \textit{select}, \textit{contrast}, or \textit{compose} interpretive lenses. This is consistent with Ric{\oe}ur's view that texts remain open to ongoing \textit{refiguration} across contexts \citep{ricoeur1976interpretation, ricoeur1991fromtext} and makes pluralization of readings a default design choice rather than an afterthought.

\vspace{-3pt}

\subsection{Evaluation Criteria for Hermeneutic Quality}
\label{subsec:hermeneutic-quality}

\vspace{-1pt}

Given the Hermeneutic Safety Loop, the central question is not only \textit{what} the model outputs, but \textit{how} an interpretive episode is carried out under an explicit horizon and evidence constraints. To make this conduct measurable in practice, we need evaluation criteria that move beyond string- or turn-level metrics and instead assess features such as faithfulness to evidence, explicit handling of ambiguity, and coverage of multiple warranted interpretations.

Standard LLM metrics such as BLEU, ROUGE, or perplexity primarily capture surface overlap and local fluency, and they often correlate only weakly with expert judgments about meaning, faithfulness, and discourse structure \citep{fabbri2021summeval, wang2023automated}. A hermeneutic perspective instead treats \textit{interpretive quality}, understood as the adequacy of an AI-supported reading of materials under a stated horizon, as a property of an \textit{AI-mediated interpretive episode} rather than of isolated strings. In this section I group episode-level quality into six dimensions:
\textit{part--whole integrity}, \textit{ambiguity sensitivity}, \textit{plurality}, \textit{evidence discipline and traceability}, \textit{quality of limitation-signaling}, and \textit{user appropriation}. The aim is not to replace standard task metrics, but to specify complementary, auditable criteria for assessing whether an LLM-mediated workflow supports responsible co-interpretation.

\vspace{-7pt}

\paragraph{Part--Whole Integrity.}

Building on the hermeneutic circle (Schleiermacher) and Gadamer's account of understanding as a movement between parts and whole, \textit{part--whole integrity} evaluates whether local claims are both \textit{faithful} (entailed or directly warranted by the evidence) and \textit{coherently integrated} into an overall interpretation under the active horizon. Operationally, this dimension has two components.
\textit{Local fidelity} requires that each checkable claim tagged \textit{supported} be entailed, or at least supported without introducing additional assumptions beyond the declared horizon, by its cited evidence span. It can be approximated by QA-style probes over source passages (construct questions whose answers are fixed by the evidence set and verify that the output preserves those answers) and by reference-free scoring that correlates with faithfulness to a given source, such as BARTScore \citep{eyal2019question, deutsch2021towards, deutsch2021understanding, yuan2021bartscore}. QA-based factual evaluation such as QAFactEval \citep{fabbri2021qafacteval} and factual error audits such as FRANK \citep{pagnoni2021understanding} can be used as claim-level deviation checks against the evidence.
\textit{Holistic integration} requires that the output preserve the discourse structure of the document as a whole: the main content units and a specified set of relations among them (for example, temporal order, causal dependence, support/attack relations, scope and exception structure) as they are warranted by the evidence under the horizon. Coverage can be assessed against authorial content units, for example Pyramid-style Summary Content Units \citep{nenkova2004evaluating}. Cross-part logical consistency can be tested via entailment and contradiction checks such as SummaC \citep{laban2022summac}, applied across the full output rather than only at the sentence level. A part--whole integrity score should therefore report (i) a local faithfulness/error rate over claims and (ii) a whole-structure preservation score over content-unit coverage and relation consistency, so that global coherence cannot be achieved simply by paraphrasing while dropping conflicts or exceptions warranted by the evidence.

\vspace{-7pt}

\paragraph{Ambiguity Sensitivity.}

In hermeneutics, ambiguity and polysemy are resources rather than defects. \textit{Ambiguity sensitivity} evaluates whether the system \textit{detects} and \textit{communicates} underdetermination: cases where the available evidence, interpreted under the declared horizon, does not warrant a single reading without introducing additional, unstated assumptions.
The system should output an \textit{ambiguity register}: a list of ambiguity points (or loci). Each locus is an auditable pointer to \textit{where} underdetermination arises, specified at the finest practically useful granularity, such as (i) a term or phrase whose sense is unclear, (ii) a referent whose identity is not fixed, (iii) a scope or quantifier relation whose range is not specified, (iv) a time frame or temporal anchor consistent with more than one reading, (v) a jurisdictional or doctrinal parameter not determined by the prompt, or (vi) an inferential step where multiple conclusions remain compatible with the evidence.
For each ambiguity locus, the system should: (i) name the locus (by quoting the relevant span, identifying the disputed parameter, or referencing the specific inferential step), so that a reviewer can locate it without further interpretation; (ii) state at least two \textit{compatible readings} at the level of claims or parameters, rather than simply asserting that ``ambiguity exists''; and (iii) handle the effects of unresolved ambiguity on later claims that depend on it. When the ambiguity depends on missing user intent or missing parameters, the system should ask a targeted disambiguating question (for example, ``Which jurisdiction applies?'', ``Which time period do you mean?'', or ``Does \textit{it} refer to X or Y?''). When the ambiguity cannot be resolved without changing the horizon or expanding the evidence set, any downstream claims that depend on resolving the locus should be explicitly labeled as \textit{conditional} on the chosen reading, and alternatives should be kept separate rather than silently collapsed to one.
Ambiguous QA benchmarks such as AmbigQA or AmbigNQ provide controlled testbeds for measuring whether ambiguity is detected and whether the system requests disambiguating information rather than defaulting to a single reading \citep{min2020ambigqa}. Calibration is required at the level of expressed confidence: when one or more unresolved ambiguity points remain, the system should either assign lower probabilities, or use qualified language (e.g., `may', `likely', `uncertain') in its claims and avoid categorical phrasing; confidence should increase only when the ambiguity register is reduced through disambiguating input or through an explicit narrowing of scope within the declared horizon, consistent with work on model self-knowledge and calibration \citep{kadavath2022language}.

\vspace{-4pt}

\paragraph{Plurality.}

\textit{Plurality} evaluates whether the system can present more than one interpretation when the declared horizon and the available evidence permit multiple warranted readings, rather than emitting a single answer by default. The system should therefore output an explicit \textit{interpretation set}: a list of alternatives, each treated as a structured object rather than as an unstructured list of qualifications.
A minimal requirement is \textit{plural rendering}: the system outputs at least two alternatives that are \textit{not} mere rephrasings. To count as genuinely distinct, alternatives must differ concretely in at least one of the following ways: (i) a conclusion that affects the user’s choice or judgment, (ii) the cited evidential basis differs (for example, the alternatives cite different supporting passages, or use the same passages to support different claims), or (iii) the governing horizon conditions differ (for example, the alternative explicitly adopts a different jurisdiction, doctrinal stance, ontology, time window, or value overlay). Each alternative should begin with a concise \textit{conditions header} (one–two sentences) that states its horizon conditions and any additional assumptions required for that alternative to be warranted. If an alternative depends on resolving an ambiguity locus, it should state the resolution it assumes.
Each alternative should also carry claim-level evidential status. Claims presented as supported must be accompanied by in-scope citations with locator precision (and quoted or highlighted spans when available). Claims that go beyond what the cited evidence entails should be tagged as \textit{extrapolation} or \textit{uncertain}. This constraint prevents plurality from collapsing into arbitrary variation: an alternative counts as a warranted interpretation only if it is either supported by in-scope evidence under a fixed horizon, or explicitly conditional on a stated horizon change or additional assumption.
Where benchmarks define multiple acceptable answers, plurality can be evaluated as coverage of the acceptable set (for example, the fraction of acceptable answers recovered), using ambiguity-oriented benchmarks such as AmbigQA or AmbigNQ when applicable \citep{min2020ambigqa}. In applied settings, alternatives that lack both in-scope evidential support and an explicit statement of the horizon they depend on should be labeled as hypotheses and should not be presented as having the same evidential status as supported interpretations.

\vspace{-4pt}

\paragraph{Evidence Discipline and Traceability.}

\textit{Evidence discipline and traceability} evaluate whether an episode renders every asserted point auditable under an explicitly declared horizon, for example, a jurisdiction, doctrinal frame, ontology, policy lens, time window, and allowed corpus. The output should be decomposed into \textit{atomic claims}: propositions small enough that a single source span is enough to judge whether they are supported.
For each atomic claim, the system should assign exactly one evidential status and expose it in the surface form of the response. A claim is either \textit{supported}, in which case it must carry one or more in-scope citations with precise location information (document identifier plus a stable location such as section heading and paragraph number, or page number and line/paragraph range, and—when the text is available—a quoted or highlighted supporting span), or it is \textit{uncertain} / \textit{unsupported}, in which case it must be explicitly marked as such and should be explicitly hedged or marked as uncertain, rather than phrased as simple assertions. If the system states a claim that goes beyond what the cited text entails, it should be tagged \textit{extrapolation} and visibly separated from supported claims (for example, in a different section or with a distinct label).
Horizon adherence is enforced by restricting admissible citations to sources within the declared horizon. Every citation attached to a \textit{supported} claim must fall within the declared horizon. If the system changes the horizon (for example, switches jurisdiction, allowed corpus, interpretive lens, or time window), it should state the new horizon \textit{before} issuing further supported claims and record the trigger for the change (for example, a user instruction or an explicit correction).
When retrieval is used, citations should refer to retrieved passages rather than to non-retrieved background knowledge. Any claim that cannot be grounded in the retrieved set should be tagged \textit{uncertain} (if the system has no basis) or \textit{extrapolation} (if it extends the retrieved evidence). Automated grounding metrics such as RAGAS can serve as secondary diagnostics of retrieval use and faithfulness, but they do not replace explicit claim-level status tags and locatable citations \citep{es2024ragas}. In practice, this discipline can be supported by cite-while-generating procedures that enforce a quote-then-answer pattern \citep{menick2022teaching} and by post-editing routines that locate and repair unsupported claims \citep{gao2022rarr}.
Each episode should maintain a \textit{provenance ledger}. For every atomic claim, the ledger records: a claim identifier; the claim text; its evidential status (\textit{supported}, \textit{uncertain}, \textit{unsupported}, or \textit{extrapolation}); the horizon identifier and version in force when the claim was produced; and the evidence locators (source identifiers plus location markers and, when available, the supporting span). This information should be sufficient for an auditor to reconstruct, for any claim, which horizon governed it and exactly which source passages were cited in support \citep{rashkin2023measuring}.

\vspace{-7pt}

\paragraph{Procedural Reflexivity.}

\textit{Procedural reflexivity} assesses whether the episode includes an explicit critique-and-revision step that identifies checkable weaknesses (evidence gaps, horizon drift, unresolved ambiguities) and repairs them in a traceable way. It does not attribute reflexive consciousness to the system; instead, it evaluates whether the workflow includes a \textit{self-audit step} that (i) identifies concrete problems that can be externally checked in the current draft under the active horizon and evidence constraints and (ii) produces revisions that measurably reduce errors that can be checked against the evidence or ground truth.
An episode should therefore expose a revision trace: an initial draft, an explicit critique of that draft, and a revised draft with updated provenance annotations. The critique must be concrete: it should identify specific claims that are likely incorrect or unsupported (for example, claims lacking citations; claims whose cited passages do not entail them; claims that conflict with other claims in the same output; claims that rely on an assumption about sources, scope, or values that is not recorded in the current horizon; or claims that require missing user parameters) and state the reason for concern in an auditable form (for example, ``no in-scope citation is attached'', ``the cited span does not support the asserted conclusion'', ``two paragraphs contradict on point X'', or ``the jurisdiction/time window is not fixed'').
Procedural reflexivity is evaluated by what changes across critique rounds. At minimum, the revised draft should show a lower rate of unsupported-claim (claims marked as supported without adequate evidence locators), internal contradiction (inconsistencies within the output), and entailment failures (claims not licensed by the cited evidence when tested with QA-style checks or entailment checks on the cited spans). Critiques should also correctly localize their targets: when a critique flags a claim as unsupported, contradictory, or horizon-misaligned, that diagnosis should be confirmable from the provenance ledger and the text. Finally, revisions should actually repair the identified problems, for example by adding appropriate citations, correcting or retracting claims, adding explicit conditionality, asking the missing disambiguating question, or declaring and justifying a necessary horizon change.
Because model rationales can be post hoc \citep{jacovi2020towards,cushman2020rationalization}, any self-critique or justification should be evidence-linked. If the critique asserts that a claim follows from a source, it should point to the specific supporting span. If it asserts that a limitation is due to missing evidence or to a horizon constraint, the corresponding gap or constraint should be visible in the declared horizon and in the provenance ledger; a mismatch between what the critique claims and what the logged evidence shows counts as a failure of self-audit, even if the final prose appears plausible.

\vspace{-3pt}

\paragraph{User Appropriation.}

Finally, \textit{user appropriation} (in a Ric{\oe}urian sense) concerns whether outputs remain usable as outputs that users can contest, revise, and integrate into their own reasoning, rather than functioning as authoritative verdicts. Since appropriation is ultimately a human act, evaluation here targets (i) interface and workflow features that make such use and modification easier and (ii) behavioral indicators of user uptake in logs.
At the interface and workflow level, the system should provide explicit controls to request alternatives, to identify which claims depend on which assumptions or horizon parameters, and to edit outputs while preserving or updating provenance (for example, edits inherit or explicitly revalidate citations and horizon tags). In logged episodes, indicators of uptake include whether users introduce substantive revisions (e.g., changing at least one claim, assumption, or horizon parameter) and whether those revisions remain traceable (i.e., the provenance ledger is updated rather than collapsed). Disagreement and uncertainty should be preservable in the record rather than smoothed away during consolidation, and final commitments in high-impact settings (e.g., clinical, legal, or safety-critical decisions) should be tied to an explicit human sign-off step with a recorded rationale (e.g., a short justification linked to the final decision).

In keeping with a hermeneutic stance, automated tools can support but do not replace human judgment. The central requirement is reasoning whose horizon, evidence use, and alternatives are explicitly logged and reviewable. Reported jointly, these six dimensions provide a basis for assessing whether an LLM-mediated workflow supports responsible co-interpretation in the sense above, or instead hides unresolved uncertainty and alternative readings.

\vspace{-1pt}

\subsection{Digital Hermeneutics as Human-Facing Literacy}
\vspace{-1pt}

As AI-generated texts are integrated into search engines, classrooms, and professional workflows, practitioners and the public alike need skills similar to those that hermeneutic philosophers have long advocated: questioning assumptions, situating texts within broader contexts, and remaining alert to bias and ambiguity. This digital hermeneutics forms the human-facing counterpart to technical work on LLM design.

% \vspace{2pt}

In much the same way that Gadamer stresses how understanding arises in an ongoing dialogue between the horizon of the text and that of the interpreter, digital hermeneutics teaches users to see AI outputs as \textit{interlocutors} rather than oracles\footnote{That is, as prompts for further questioning rather than final answers.}. LLM-generated content often sounds authoritative, yet it rests on statistical pattern matching over historical corpora, without the lived experience and horizon or dialogical engagement that Gadamer considers vital for genuine understanding. A hermeneutic outlook therefore treats AI-generated statements as \textit{proposals} that require independent investigation and critical appraisal rather than as fully formed truths \citep{pinell2024does, floridi2023ai}.

This training has several elements. First, users should learn to contextualize AI outputs. An LLM's text is detached from any single human authorial horizon and instead emerges from a vast and impersonal training corpus. Exercises might involve comparing LLM responses with authoritative domain sources or highlighting cultural and historical gaps in the model's output \citep{maciag2024hermeneutics}. Second, they should learn to question apparent neutrality. Drawing on Ric{\oe}ur's hermeneutics of suspicion, students and professionals can ask whose voice a given answer represents and whose perspectives are missing. In a Derridean spirit, they can remain alert to how textual meaning is shaped by differences in language, culture, and power, and by training data biases \citep{derrida1978writing, vromen2024language}. Third, they should practice engaging with the system through repeated back-and-forth questions and answers. In line with Gadamerian dialogue, users can be encouraged to pose follow-up questions, explore alternative interpretations, and prompt the model to reframe its answers from new angles. This iterative engagement strengthens critical analysis and reveals ambiguities that a single authoritative-sounding response would mask \citep{demichelis2024hermeneutic, delacroix2025designing, delacroix2025moral}. Fourth, users in domains such as medicine, law, and scholarship should learn to attend to ethical and disciplinary specifics. For instance, clinicians can be trained to cross-check AI-generated recommendations against existing guidelines and ethical principles so that discrepancies between AI output, professional judgment, and the patient's situation can be noticed and corrected, rather than silently absorbed.

% \vspace{2pt}
 
Some universities have begun formalizing this approach through digital hermeneutics modules, where students critically analyze AI outputs alongside traditional texts and learn to situate both within their historical and social contexts \citep{vanNuenen2023}. In professional settings, similar guidelines can embed these interpretive skills into routine tasks such as verifying the provenance of AI-generated content or seeking out the perspectives of groups that are historically marginalized in training data.

% \vspace{2pt}

Ultimately, digital hermeneutics reminds us that the user, not the model, bears interpretive responsibility. Just as Schleiermacher and Heidegger emphasize the inescapable role of the interpreter's prior understandings, so too we must acknowledge our own prejudices, in Gadamer's sense of the term, when engaging with LLMs. Rather than accepting AI outputs uncritically, this literacy cultivates a reflective, question-driven way of interacting with AI and helps ensure that human interpretive agency remains at the center of AI-mediated meaning-making.

\vspace{-4pt}

\section{Hermeneutic Quality and Societal Dynamics}
\label{sec:societal-dynamics}
\vspace{-2pt}

AI \& SOCIETY scholarship is concerned not only with conceptual distinctions, but also with the ways AI technologies alter existing social practices and generate new ones. Hermeneutic quality---the extent to which AI-mediated interpretation respects plurality, historicity, and part--whole integrity---matters across several interpretive domains. In what follows, I focus on three such domains and, in each case, sketch what current LLM use tends to miss, why these omissions matter socially, and how hermeneutic design can make interpretive practices more plural, traceable, and accountable.

\vspace{-4pt}

\subsection{Law and Public Policy}
\vspace{-1pt}

Legal practice is saturated with interpretation: of statutes, contracts, precedents, and constitutional principles. Law firms are already using LLMs to summarize cases, draft memoranda, and help develop arguments, including in response to opposing counsel who uses similar tools. The familiar risks include hallucinated citations, mischaracterized holdings, and confusion about jurisdiction or temporal scope. Here we highlight a further risk we call hermeneutic flattening: models can smooth over the controversies, disagreements, and ambiguities that structure legal interpretation, presenting contested matters as if they were settled legal truths.

From the standpoint of hermeneutic quality, current practice typically lacks explicit jurisdictional and doctrinal horizons, systematic presentation of alternative lines of interpretation (for example different schools of statutory interpretation or competing readings of a precedent), and clear evidence discipline that ties model claims to the passages and precedents said to support them. A one-shot answer with vague sourcing and no indication of doctrinal alternatives encourages automation bias and encourages users to treat a single output as the legal view.

A legal LLM assistant should operate within an explicit artificial horizon when used for professional legal work. For a given matter this horizon would state, in plain language: (i) which jurisdiction and court levels are in scope; (ii) which time window of case law and statutes is considered; (iii) which approach to interpretation is used as default (for example, textualist, purposive, or originalist); and (iv) which secondary sources and treatises are treated as authoritative. This statement appears to the user at the top of the interaction and is stored in the case file. The system labels its outputs accordingly: it marks a passage as a textualist reading or as a purposive reading and records which horizon it used. If the user switches horizon, for example, from federal to state law, or from a textualist to a purposive stance, the interface shows that change and its time, and subsequent answers are generated under the new horizon. Citations and quotations are tied to the active horizon, so that a lawyer or judge can check whether a suggested interpretation is supported by sources that actually belong to the stated scope. In the Hermeneutic Safety Loop, choosing between competing interpretations then appears as a human decision about which horizon and which reading to endorse, not as an opaque internal choice of the model.

Socially, the stakes here are at least twofold. First, a real part of what makes law---and judicial decision-making in particular---legitimate is that citizens can see why decisions are made. Legal outcomes are supposed to be grounded in reasons that can be spelled out, inspected, and challenged, and that make sense in light of precedent and legal doctrine. If LLMs start shaping legal reasoning in ways that people cannot follow, question, or contest, then that legitimacy is put at risk. Second, hermeneutic flattening changes whose interests are actually heard and taken seriously. When AI turns complex legal disputes into streamlined summaries, it can also streamline away perspectives that are already easy to miss. If minority or marginalized voices repeatedly drop out of AI-mediated explanations—whether in internal case workflows or in the versions that circulate publicly through different media—then whole groups may find their concerns becoming less visible in legal discourse, even if the formal doctrinal record still acknowledges them. For that reason, hermeneutically informed legal tools should aim for more than ``more accurate'' summaries. They should support workflows where human lawyers remain clearly responsible as the final interpreters, and where AI helps broaden the range of legal reasons that get articulated and weighed, rather than quietly narrowing it.

\vspace{-3pt}

\subsection{Education and Scholarly Knowledge}

\vspace{-1pt}

In educational and scholarly contexts, LLMs are used to explain concepts, generate study materials, and summarize articles and books. They can substantially reduce barriers to basic explanatory material, for example by making basic explanations and summaries available to people without institutional library access or expert supervision. At the same time, they can produce a form of superficial understanding: statements that look correct in isolation but that are de-contextualized and that obscure the interpretive struggles, methodological disagreements, and historical developments through which a field has reached its present vocabulary.

In everyday use, LLM-generated explanations often lack explicit disciplinary horizons, such as the schools, paradigms, and methodological commitments that structure a field. They tend to present a blended summary as if it were neutral, with little indication of the plurality of scholarly positions on contested questions or of the historical sequence in which ideas emerged. Standard interfaces do not, by default, provide mechanisms that require students to appropriate outputs critically rather than copying them. A student can easily accept a polished explanation without seeing which parts are controversial, which are simplifications, and which reflect specific traditions within the discipline.

Design informed by hermeneutics, together with training in digital hermeneutics, can make these horizons and differences explicit in everyday use. Educational applications can prompt models to introduce canonical alternatives—for example, by indicating that a topic is debated between two or three major positions and briefly stating how they differ. Interfaces can encourage users to request contrastive explanations: for a phenomenon, a hermeneutic account can be paired with the positivist alternative (and vice versa), while a continental treatment of a problem can be paired with the analytic alternative (and vice versa). Assignments can require students to critique, revise, and check AI-generated explanations of texts against the source and against scholarly reviews or summaries, treating model outputs as provisional study notes rather than definitive summaries.

The societal implications concern the formation of intellectual autonomy. If a generation learns primarily from AI-generated summaries rather than engaging with the analysis and interpretation found in scholarship, public discourse may rely more on unexamined shorthand and become less able to deal with disagreement. The relevant hermeneutic point is that educational practices can cultivate habits of reading, questioning, contextualizing, and comparing—skills needed for participation in public discussion. Conversely, LLMs can be used educationally in ways that foster these same habits, supporting democratic citizenship and more responsible engagement with expertise.

\subsection{Moral, Religious, and Cultural Discourse}

Religious exegesis, moral reasoning, and the interpretation of historical events are domains in which interpretive disagreements are deep, often identity-defining, and closely tied to communal traditions. Here too LLMs are already treated by some users as neutral or quasi-neutral arbiters. Users ask what a verse really means, request a summary of a moral debate, or seek an explanation of the causes of a conflict. The interaction pattern is typically configured to elicit a single, apparently balanced answer.

In these settings, the illusion of understanding is particularly salient here. Model outputs may reflect majority or dominant traditions while marginalizing minority theologies, ethical systems, or historical memories. They may present normative positions as descriptive facts, for example by summarizing a controversial teaching as if it were universally accepted within a religion, or they may project contemporary assumptions back onto texts and events from very different horizons (historical periods and cultural settings). Because the model has no lived relation to the traditions it synthesizes, the responsibility for recognizing these distortions falls entirely on users and communities.

Here the notion of an artificial horizon applies directly. A scriptural assistant, for example, should declare at the outset which tradition and corpus it is using: it might state that it is operating within a specific denomination, that it relies on a named set of commentaries, and that it assumes a particular hermeneutic stance on the status of the text. That declaration is the artificial horizon for the session. The assistant then generates exegesis only under that horizon and labels its answers accordingly. Users can request a different horizon, for instance, a historical-critical reading or the stance of another denomination, and the system switches to a second, clearly labeled horizon so that the two readings can be compared side by side. In political and historical applications, a similar structure would require the system to identify the standpoint from which it explains a conflict, to point to sources that support that standpoint, and to offer at least one alternative horizon that incorporates different sources or affected groups. In all these cases, the horizon is not an invisible default but a configurable and documented object. Horizon engineering (designing and governing horizons) in moral and religious domains therefore consists in defining which horizons are treated as authorized or appropriate within a given community, exposing them to users, and keeping human authorities responsible for which horizons are available and how they are changed.

Public moral and cultural discourse is not only about information; it also concerns whose voices count, which traditions are taken seriously, and how disagreement is handled in practice. In this sense, ethics of communication is socially consequential because communicative practices shape representation, convey respect for different cultural, social, and religious forms of life, and structure the terms of contestation. LLMs that meet criteria of hermeneutic quality can help systematize and surface a wider range of possible meanings, making room for alternative voices and explicit dispute. LLMs that fall short may instead narrow interpretive space, reinforce dominant narratives, and present their partiality under a veneer of neutral fluency, as if embedded biases were objective or neutral.

\section{Conclusion}

LLMs are employed for interpretation when interpretation is understood as the activity of bringing to understanding what is taken to matter---whether fact or value. On this view, interpretation involves rendering the facticity of something intelligible in much the same way that we render a text intelligible as meaningful. I argue that an LLM's output should be treated as a fallible interpretive suggestion within an AI-mediated interpretive cycle, in which only humans possess historical situatedness and, with it, agency.
I therefore distinguish between hermeneutic understanding of a phenomenon and artificial or algorithmic interpretation. This distinction clarifies how LLMs can be described and designed for interpretive purposes without attributing artificial agency to the model, while also reusing established interpretive techniques that support hermeneutic quality. I map these techniques to design patterns that, among other things, facilitate plurality, make horizons explicit, discipline the use of evidence, and enable user appropriation of technology. These patterns---grounded in traceable episodes of interpretation---serve as a reference point for discussing LLM-mediated interpretation across domains with significant stakes, including law, education, and public moral and cultural discourse. In these settings, poor hermeneutic quality may undermine legitimacy, the adequate socialization of youth for civic participation, and the representation of minority traditions and perspectives.
Across such cases, responsible employment of LLMs depends less on increasing technological autonomy than on designing systems and practices that preserve human interpretive responsibility and the plural, contestatory character of the hermeneutic process.

\section*{Acknowledgment}
This work was supported by the Swiss National Science Foundation (SNSF) under Grant No.~222339.  
The author would like to thank Dr.~Habiballah Rahimi Eichi, who first introduced the author to the philosophy of hermeneutics.

\section*{Statements and Declarations}
AI tools were used for language polishing---to correct grammar, improve clarity, and standardize phrasing in the English text. They were not used for ideation, conceptualization, literature review, and citation generation. The arguments, structure, citations, and final prose were composed and verified by the author, who takes full responsibility for the content. No generative-AI images were used. In this limited sense, the paper is an instance of human--AI co-interpretation for responsible AI: AI was used only to propose wording edits, and all changes were subject to the author's review and final approval.

% \printbibliography
% ========================================= 
%              References
% ========================================= 
% \bibliographystyle{IEEEtran}
\bibliographystyle{tmlr}
\bibliography{references}

\end{document}

%% file: preamble.tex
\usepackage[T1]{fontenc}
\usepackage[utf8]{inputenc}
\DeclareUnicodeCharacter{0304}{} % suppress combining macron if it appears in pasted text

\usepackage{microtype}

\usepackage{geometry}
\usepackage{amsmath,amssymb,amsfonts,mathtools}

\usepackage{graphicx}
\usepackage{booktabs}
\usepackage{tabularx}
\usepackage{array}
\usepackage{ragged2e}

\newcolumntype{L}[1]{>{\raggedright\arraybackslash}p{#1}}
\newcolumntype{J}[1]{>{\justifying\arraybackslash}p{#1}}

\usepackage{enumitem}

\usepackage{tikz}
\usetikzlibrary{arrows.meta,calc,fit,positioning,shapes,shapes.geometric}

\usepackage{hyperref}
\hypersetup{
  colorlinks = true,
  linkcolor  = red!50!black,
  citecolor  = blue!50!black,
  urlcolor   = black
}

\usepackage{comment}